\documentclass[runningheads]{llncs}
\usepackage{graphicx}
\usepackage{amsmath,amssymb} 
\usepackage{color}
\usepackage[dvipsnames,table]{xcolor}
\usepackage{multirow}
\usepackage{array}
\usepackage{caption}
\usepackage{subcaption}
\usepackage{csquotes}
\usepackage[font=footnotesize]{caption}
\usepackage{soul} 
\usepackage{hyperref}
\usepackage{pdfpages}

\graphicspath{{Images/}}
\DeclareMathOperator*{\argmax}{arg\,max}
\newcolumntype{P}[1]{>{\centering\arraybackslash}p{#1}}

\renewcommand{\footnotesize}{\scriptsize}

\begin{document}
\title{Sidekick Policy Learning \\ for Active Visual Exploration} 

\titlerunning{Sidekick Policy Learning for Active Visual Exploration}
%
\author{Santhosh K. Ramakrishnan\inst{1} \and Kristen Grauman\inst{2}}
%
\authorrunning{S. Ramakrishnan and K. Grauman}
%
\institute{The University of Texas at Austin, Austin, TX 78712 \and
           Facebook AI Research, 300 W. Sixth St. Austin, TX 78701 \\
           \email{srama@cs.utexas.edu, grauman@fb.com\thanks{\it{On leave from University of Texas at Austin (grauman@cs.utexas.edu)}}}}

%
\maketitle              
\begin{abstract}
We consider an \emph{active visual exploration} scenario, where an agent must intelligently select its camera motions to efficiently reconstruct the full environment from only a limited set of narrow field-of-view glimpses.
  While the agent has full observability of the environment during training, it has only partial observability once deployed, being constrained by what portions it has seen and what camera motions are permissible. We introduce \emph{sidekick policy learning} to capitalize on this imbalance of observability.  The main idea is a preparatory learning phase that attempts simplified versions of the eventual exploration task, then guides the agent via reward shaping or initial policy supervision.  To support interpretation of the resulting policies, we also develop a novel policy visualization technique. Results on active visual exploration tasks with $360^{\circ}$ scenes and 3D objects show that sidekicks consistently improve performance and convergence rates over existing methods. Code, data and demos are available~\footnote{Project website: \url{http://vision.cs.utexas.edu/projects/sidekicks/}}.

\end{abstract}
\section{Introduction}

Visual recognition has witnessed dramatic successes in recent years.  Fueled by benchmarks composed of Web photos, the focus has been inferring semantic labels from \emph{human-captured images}---whether classifying scenes, detecting objects, or recognizing activities~\cite{ILSVRC15,lin2014microsoft,soomro2012ucf101}.  By relying on human-taken images, the common assumption is that an intelligent agent will have already decided where and how to capture the input views.  While sufficient for handling static repositories of photos (e.g., auto-tagging Web photos and videos), assuming informative observations glosses over a very real hurdle for embodied vision systems.

A resurgence of interest in perception tied to action takes aim at that hurdle.   In particular, recent work explores agents that optimize their physical movements to achieve a specific perception goal, e.g., for active recognition~\cite{germs-bmvc2015,dinesh-eccv2016,johns-cvpr2016,ammirato-icra2017,dinesh-pami2018}, visual exploration~\cite{dinesh-ltla}, object manipulation~\cite{levine2016handeye,pinto2016supersizing,nair2017combining}, or navigation~\cite{zhu-iccv2017,gupta2017unifying,ammirato-icra2017}.
In any such setting, deep reinforcement learning (RL) is a promising approach.  The goal is to learn a policy that dictates the best action for the given state, thereby integrating sequential control decisions with visual perception.

\begin{figure}[t!]
    \centering
      \includegraphics[width=1\textwidth, clip, trim={0.1cm 9.2cm 8cm 0cm}]{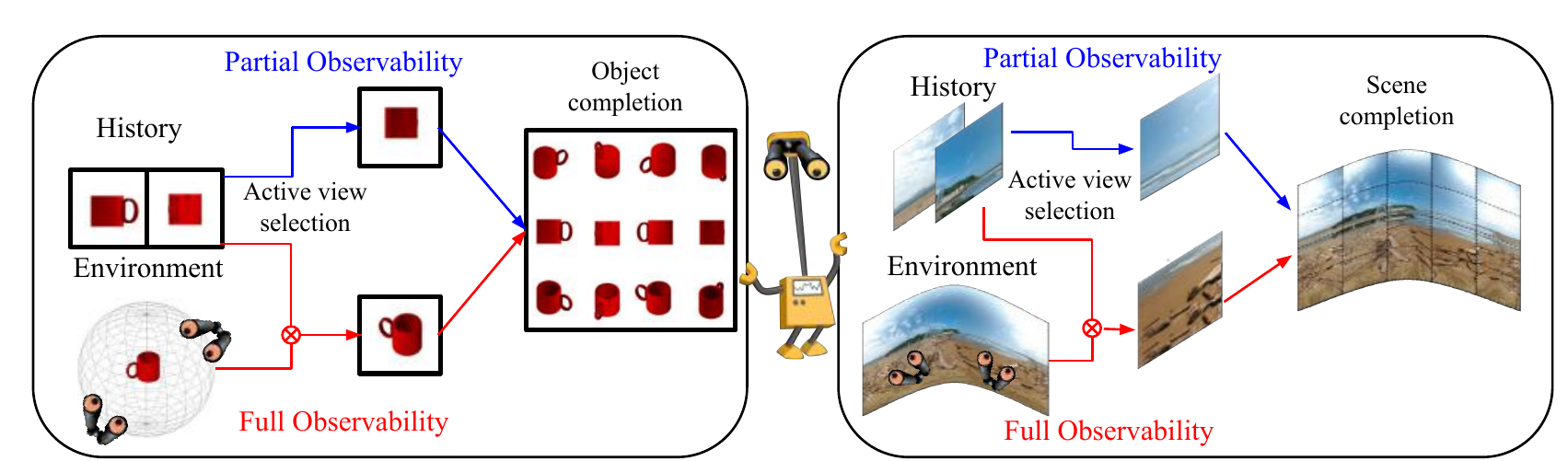}
      \caption{\footnotesize Embodied agents that actively explore novel objects (left) or $360^{\circ}$ environments (right)
intelligently select camera motions to gain as much information as possible with very few glimpses.  While they naturally face \textcolor{Blue}{limited observability} of the environment, during \emph{learning} \textcolor{Red}{fuller observability} may be available.  We propose sidekicks to guide policy learning for active visual exploration.} 
  \vspace*{-0.15in}
    \label{examples}
\end{figure}

However, costly exploration stages and partial state observability are 
well-known impediments to RL.
In particular, an active visual agent~\cite{zhu-iccv2017,gupta2017unifying,zhu2017target,dinesh-ltla} has to take a long series of actions purely based on the limited information available from its first person view. Due to poor action selection based on limited information, the most effective viewpoint trajectories are buried among many mediocre ones, impeding the agent's exploration in complex state-action spaces.

We observe that agents lacking full observability when deployed may nonetheless possess full observability \emph{during training}, in some cases.  Overall, the imbalance occurs naturally when an agent is trained with a broader array of sensors than available at test-time, or trained free of the hard time pressures that limit test-time exploration.  In particular, as  we will examine in this work, once deployed, an active exploration agent can only move the camera to ``look-around" nearby~\cite{dinesh-ltla}, yet if trained with omnidirectional panoramas, could access any possible viewpoint while learning.  Similarly, an active object recognition system~\cite{dinesh-eccv2016,johns-cvpr2016,ammirato-icra2017,shapenet,dinesh-pami2018} can only see its previously selected views of the object; yet if trained with CAD models, it could observe all possible views while learning. Additionally, agents can have access to  multiple sensors during training in simulation environments~\cite{dosovitskiy2017carla,pinto2017asymmetric,embodiedqa}, yet operate on first-person observations during test-time.
However, existing methods restrict the agent to the same partial observability during training~\cite{shapenet,johns-cvpr2016,dinesh-eccv2016,dinesh-ltla,zhu-iccv2017,dinesh-pami2018}.

We propose to leverage the imbalance of observability.  
To this end, we introduce \emph{sidekick policy learning}.  We use the name ``sidekick" to signify how a sidekick to a hero (e.g., in a comic or movie) provides alternate points of view, knowledge, and skills that the hero does not have.  In contrast to an \emph{expert}~\cite{guo2014deep,vapnik2016learning}, 
a sidekick \emph{complements} the hero (agent), yet 
cannot solve the main task at hand.

We propose two sidekick variants.  Both use access to the full state during a preparatory training  period to facilitate the agent's ultimate learning task.  The first sidekick previews individual states, estimates their value, and shapes rewards to the agent for visiting valuable states during training.  The second sidekick provides initial supervision via trajectory selections to accelerate the agent's training, while gradually permitting the agent to act on its own.  In both cases, the sidekicks learn to solve \emph{simplified} versions of the main task with full observability, and use insights from those solutions to aid the training of the agent.  At test time, the agent has to act without the sidekick. 

We validate sidekick policy learning for active visual exploration~\cite{dinesh-ltla}.  The agent enters a novel environment and must select a  sequence of camera motions to rapidly understand its entire surroundings.
For example, an agent that has explored various grocery stores should enter a \emph{new} one and, with a couple glimpses, 1) conjure a belief state for where different objects are located, then 2) direct its camera to flesh out the harder-to-predict objects and contexts.  
The task is like active recognition~\cite{shapenet,johns-cvpr2016,dinesh-eccv2016,ammirato-icra2017}, except that the training signal is \emph{pixelwise} reconstruction error for the full environment rather than labeling error.
Our sidekicks can look at any part of the environment in any sequence during training, whereas the actual agent is limited to physically feasible camera motions and sees only those views it has selected.  On two standard datasets~\cite{xiao2012recognizing,shapenet}, we show how 
sidekicks  accelerate training and promote better look around policies.  


As a secondary contribution, we present a novel policy visualization technique.  Our approach takes the learned policy as input, and displays a sequence of heatmaps showing regions of the environment most responsible for the agent's selected actions.  The resulting visualizations help illustrate how sidekick policy learning differs from traditional training.

\vspace*{-0.1in}
\section{Related Work}
\vspace*{-0.05in}


\noindent\textbf{Active vision and attention:} 
Linking intelligent control strategies to perception has early foundations in the field~\cite{aloimonos1988active,ballard1991animate,bajcsy1988active,wilkes1992active}.  Recent work explores new strategies for active object recognition~\cite{shapenet,johns-cvpr2016,dinesh-eccv2016,ammirato-icra2017,dinesh-pami2018}, object localization~\cite{caicedo2015active,gupta2017cognitive,zhu2017target}, and visual SLAM~\cite{kim2013perception,spica2014active}, in order to minimize the number of sampled views required to perform accurate recognition or reconstruction.
Our work is complementary to any of the above: sidekick policy learning is a means to accelerate and improve active perception when observability is greater during training.

Models of saliency and attention allow a system to prioritize portions of its observation to reduce clutter or save computation~\cite{liu2011learning,ba2014multiple,mnih2014recurrent,yang2013saliency,xu2015show}.  However, unlike both our work and the active methods above, they assume full observability at test time, selecting among already-observed regions.
Work in active sensor placement aims to place sensors in an environment to maximize \emph{coverage}~\cite{dhillon2003sensor,krause2007near,wang2011coverage}.  
We introduce a model for coverage in our policy learning solution (Sec.~\ref{sec:demo}).  However, rather than place and fix $N$ static sensors, the visual exploration tasks entail selecting new observations dynamically and in sequence.

\vspace*{0.05in}
\noindent\textbf{Supervised learning with observability imbalance:}
Prior work in supervised learning investigates ways to leverage greater observability during training, despite more limited observability during test time.  Methods for depth estimation~\cite{gupta2016cross,garg2016unsupervised,tulsiani2017multi} and/or semantic segmentation~\cite{song2016im2pano3d,hong2015decoupled,hong2016learning} use RGBD depth data, multiple views, and/or auxiliary annotations during training, then proceed with single image observations at test time. Similarly, self-supervised losses~\cite{mirowski2016learning,jaderberg2016reinforcement} based on auxiliary prediction tasks at training time have been used to aid representation learning for control tasks.
Knowledge distillation~\cite{hinton2015distillation} lets a ``teacher" network guide a ``student" with the motivaton of network compression. 
In learning with privileged information, an ``expert" provides the student with training data having extra information (unavailable during testing)~\cite{vapnik2016learning,sharmanska2013learning,lapin2014learning}.
At a high level, all the above methods relate to ours in that a simpler learning task facilitates a harder one.  However, in strong contrast, they tackle supervised classification/regression/representation learning, whereas our goal is to learn a \emph{policy} for selecting actions. 
Accordingly, we develop a very different strategy---introducing rewards and trajectory suggestions---rather than auxiliary labels/modalities. \\ 

\noindent\textbf{Guiding policy learning:}
There is a wide body of work aimed at addressing sparse rewards and partial observability. 
Several works explore \emph{reward shaping} motivated by different factors.
The intrinsic motivation literature develops parallel reward mechanisms, e.g., based on surprise~\cite{pathakICMl17curiosity,bellemare2016unifying}, to direct exploration.  The TAMER framework~\cite{knox2009interactively,knox2010combining,knox2012reinforcement} utilizes expert human rewards about the end-task. Potential-based reward shaping~\cite{harutyunyan2015expressing} incorporates expert knowledge grounded in potential functions to ensure policy invariance. Others convert control tasks into supervised measurement prediction task by defining goals and rewards as functions of measurements~\cite{dosovitskiy2016learning}. 
In contrast to all these approaches, our sidekicks exploit the observability difference between training and testing to transfer knowledge from a simpler version of the task. This external knowledge directly impacts the final policy learned by augmenting task related knowledge via reward shaping.  

\emph{Behavior cloning} provides expert-generated trajectories as supervised (state, action) pairs~\cite{bojarski2016end,giusti2016machine,duan2017one,ross2011reduction}. 
\emph{Offline planning}, e.g., with tree search, is another way to prepare good training episodes by investing substantial computation offline~\cite{guo2014deep,anthony2017thinking,silver2017mastering}, but observability is assumed to be the same between training and testing.
\emph{Guided policy search} uses importance sampling to optimize trajectories within high-reward regions~\cite{levine2013guided} and can utilize full observability~\cite{levine2016end}, yet transfers from an expert in a purely supervised fashion.
Our second sidekick also demonstrates good action sequences, but we specifically account for 
the observability imbalance by annealing supervision over time. 

More closely related to our goal is the \emph{asymmetric actor critic}, which leverages synthetic images to train a robot to pick/push an object~\cite{pinto2017asymmetric}.  
Full state information from the graphics engine is exploited to better train the critic. 
While this approach modifies the advantage expected for a state like our first sidekick, this is only done at the task level. Our sidekick injects a different perspective by solving simpler versions of the task, leading to better performance (Sec.~\ref{sec:results}).

\vspace*{0.02in}
\noindent\textbf{Policy visualization:}
Methods for post-hoc explanation of deep networks are gaining attention due to their complexity and limited interpretability.  In supervised learning, heatmaps indicating regions of an image most responsible for a decision are generated via backprop of the gradient for a class label~\cite{simonyan-iclr2014,fong-vedaldi-interpretable,batra-gradcam}.
In reinforcement learning, policies for visual tasks (like Atari) are visualized using t-SNE maps~\cite{zahavy} or heatmaps highlighting the parts of a \emph{current} observation that are important for selecting an action~\cite{DBLP:journals/corr/abs-1711-00138}.  We introduce a policy visualization method that reflects the influence of an agent's \emph{cumulative} observations on its action choices, and use it to illuminate the role of sidekicks.

\section{Approach}
\vspace*{-0.05in}

Our goal is to learn a policy for controlling an agent's camera motions such that it can explore novel environments and objects efficiently.  Our key insight is to facilitate policy learning via sidekicks that exploit 1) full observability and 2) unlimited time steps to solve a simpler problem in a preparatory training phase.

We first formalize the problem setup in Sec.~\ref{sec:setup}.  After overviewing observation completion as a means of active exploration in 
Sec.~\ref{sec:completion}, 
we introduce our sidekick learning framework in Sec.~\ref{sec:sidekick}.  
We tie together the observation completion and sidekick components with the overall learning objective in 
Sec.~\ref{sec:obj}.  Finally, we present our policy visualization 
technique in Sec.~\ref{sec:vis}.

\subsection{Problem setup: active visual exploration}
\label{sec:setup}

The problem setting builds on the ``learning to look around" challenge introduced in~\cite{dinesh-ltla}. Formally, the task is as follows.  
The agent starts by looking at a novel environment (or object)
 $X$ from some unknown viewpoint~\footnote{For simplicity of presentation, we 
 represent an \emph{environment} as $X$ where the agent explores a novel scene, looking outward in new viewing directions.  However, experiments will also use $X$ as an \emph{object} where the agent moves around an object, looking inward at it from new viewing angles.}.  It has a budget $T$ of time to explore the environment.  The learning objective is to minimize the error in the agent's 
 pixelwise reconstruction of the full---mostly unobserved---environment using only the sequence of views selected within that budget.
 
Following~\cite{dinesh-ltla}, we discretize the environment into a set of candidate viewpoints.
In particular, the space of viewpoints is a
viewgrid indexed by $N$ elevations and $M$ azimuths, denoted by $V(X) = \{x(X, \theta^{(i)}) | 1 \le i \le MN \}$,
where $x(X, \theta^{(i)})$ is the 2D view of $X$ from viewpoint $\theta^{(i)}$, which is comprised of two angles.   More generally, $\theta^{(i)}$ could capture both camera angle and position; however, to best exploit existing datasets, we limit camera motions to rotations.

The agent expends the budget in discrete increments, called ``glimpses", by selecting $T-1$ camera motions in sequence. 
At each time step, the agent gets observation $x_{t}$ from the current viewpoint. The agent makes an exploratory rotation ($\delta_{t}$) based on its policy $\pi$. When the agent executes
action $\delta_{t} \in \mathcal{A}$, the viewpoint changes according to $\theta_{t+1} = \theta_{t} + \delta_{t}$.  
For each camera motion $\delta_{t}$
executed by the agent, a reward $r_{t}$ is provided by the environment (Sec.~\ref{sec:reward} and~\ref{sec:obj}).  
Using the view $x_{t}$, the agent updates its internal representation
of the environment, denoted $\hat{V}(X)$. 
Because camera motions are restricted to have proximity to the current camera angle (Sec.~\ref{sec:dataset}) and candidate viewpoints partially overlap, the discretization promotes efficiency without neglecting the physical realities of the problem (following~\cite{germs-bmvc2015,dinesh-eccv2016,dinesh-ltla,johns-cvpr2016}).

\vspace*{-0.1in}
\subsection{Recurrent observation completion network}
\label{sec:completion}

We start with the deep RL neural network architecture proposed in~\cite{dinesh-ltla} to represent the agent's recurrent observation completion.  The process is deemed ``completion" because the agent strives to hallucinate portions of the environment it has not yet seen.
It consists of five modules: \textsc{Sense}, \textsc{Fuse}, \textsc{Aggregate}, \textsc{Decode}, and \textsc{Act} with parameters $W_{s}$, $W_{f}$, $W_{r}$, $W_{d}$ and $W_{a}$ respectively.
\begin{itemize}
  \item{\textsc{Sense}: Independently encodes the view ($x_{t}$) and proprioception ($p_{t}$) consisting of elevation at time $t$ and relative motion from time $t-1$ to $t$, and returns the encoded tuple 
  $s_{t} = \textsc{Sense}(x_{t}, p_{t})$.}\vspace{0.25em}  
  \item{\textsc{Fuse}: Consists of fully connected layers that jointly encode the tuple $s_{t}$ and output a fused representation $f_{t} = \textsc{Fuse}(s_{t})$.}\vspace{0.25em}

  \item{\textsc{Aggregate}: An LSTM that aggregates fused inputs over time to build the agent's internal representation
    $a_{t} = \textsc{Aggregate}(f_{1}, f_{2}, ..., f_{t})$ of $X$. }\vspace{0.25em}  
  \item{\textsc{Decode}: A convolutional decoder which reconstructs the viewgrid\hfill\break
  $\hat{V}_{t} =\textsc{Decode }(a_{t})$ as a set of $MN$ feature maps ($3MN$ for 3 channeled
  images) corresponding to each view of the viewgrid.}\vspace{0.25em}  
  
  \item{\textsc{Act}: Given the aggregated state $a_{t}$ and proprioception $p_t$, the \textsc{Act} module outputs a probability
  distribution $\pi(\delta | a_{t})$ over the candidate camera motions $\delta \in \mathcal{A}$. 
  An action sampled from this distribution $\delta_{t} = \textsc{Act}(a_{t},p_t)$ is executed.}
\end{itemize}

\begin{figure}[t!]
    \centering
    \includegraphics[width=\textwidth, clip, trim={0cm 0cm 0cm 0cm}]{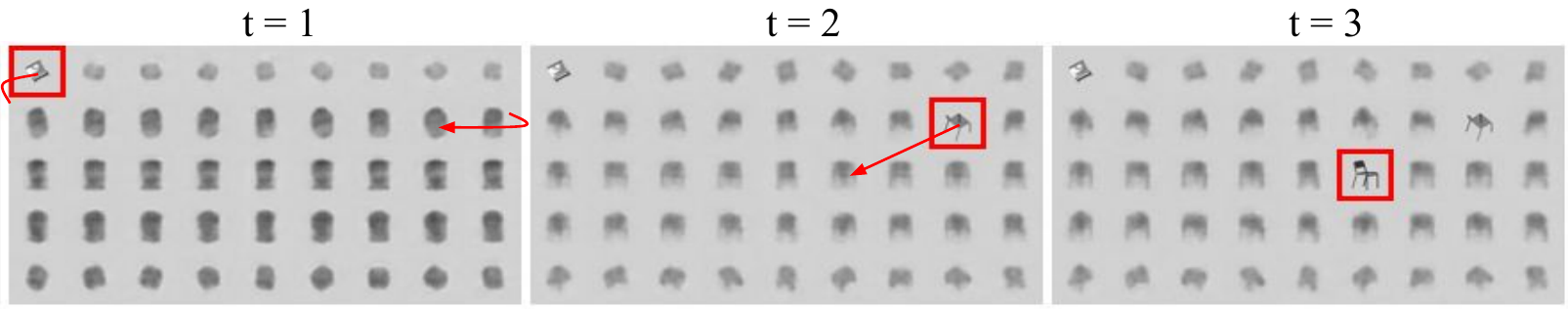}\vspace{-0.3cm}
    \caption{\footnotesize Active observation completion.  The agent receives one view (shown in red), updates its belief and reconstructs the viewgrid at each time step. It executes an action (red arrows) according to its policy to obtain the next view.  The active agent must rapidly refine its belief with well-chosen views.}
    \label{look-around-concept}
    \vspace{-0.5cm}
\end{figure}

At each time step, the agent receives and encodes a new view $x_{t}$, then updates its
internal representation $a_t$ by sensing, fusing, and aggregating. It decodes the viewgrid $\hat{V}_{t}$ and
executes $\delta_{t}$ to change the viewpoint. It repeats the above steps until the
time budget $T$ is reached (see Fig.~\ref{look-around-concept}).  See Supp.~for implementation details and architecture diagram.  
\vspace{-0.1cm}

\subsection{Sidekick definitions}
\label{sec:sidekick}

Sidekicks provide a preparatory learning phase that informs policy learning.  Sidekicks have full observability during training: in particular, they can observe the results of arbitrary camera motions in arbitrary sequence.  This is impossible for the actual look-around agent---who must enter novel environments and respect physical camera motion and budget constraints---but it \emph{is} practical for the sidekick with fully observed training samples (e.g., a $360^{\circ}$ panoramic image or 3D object model, cf.~Sec.~\ref{sec:dataset}).
Sidekicks are trained to solve a simpler problem with relevance to the ultimate look-around agent, serving to accelerate training and help the agent converge to better policies.  
In the following, we define two sidekick variants: a reward-based sidekick and a demonstration-based sidekick.

\vspace*{-0.15in}
\subsubsection{Reward-based sidekick}
\label{sec:reward}

The reward-based sidekick aims to identify a set of $K$ views 
$\{x(X,\theta_{1}), \ldots, x(X,\theta_{K})\}$ which can provide
maximal information about the environment $X$. 
The sidekick is allowed to access $X$ and select views
without any restrictions. 
Hence, it addresses a simplified
completion problem.  
 
\begin{figure}[t!]
    \centering
      \includegraphics[width=\textwidth, clip, trim={0cm 6.1cm 1.03cm 0cm}]{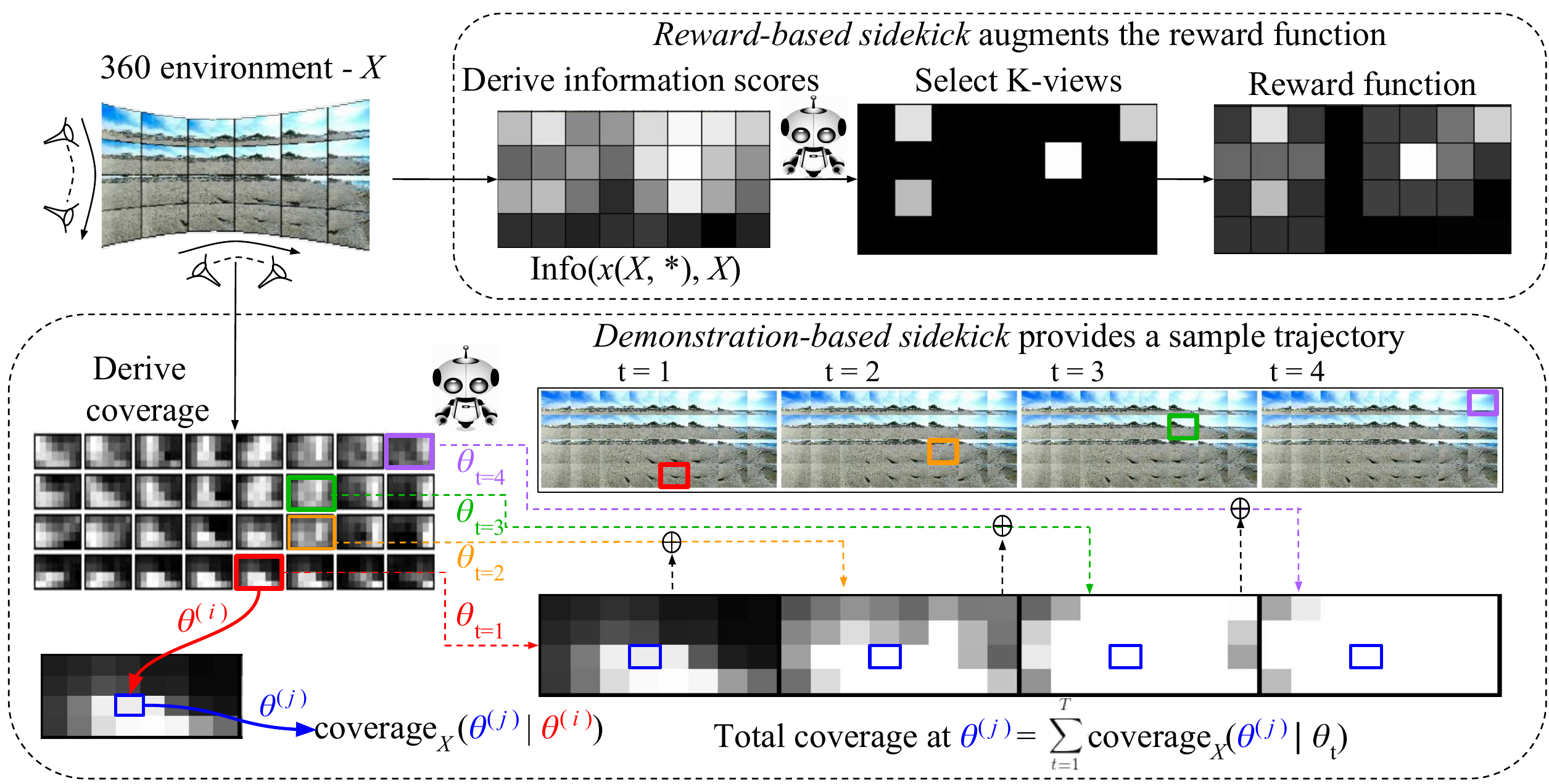}

    \caption{\footnotesize Top left shows the $360^{\circ}$ environment's viewgrid, indexed by viewing elevation and azimuth.
    \textbf{Top: Reward sidekick} 
 scores individual views based on how well they alone permit inference of the viewgrid $X$ (Eq~\ref{one_view_score}).  The grid of scores (center) is post-processed with non-max suppression to prioritize $K$ non-redundant views (right), then is used to shape the agent's rewards.
    \textbf{Bottom: Demonstration sidekick.}  Left ``grid-of-grids" displays example coverage score maps (Eq~\ref{coverage}) for all $\theta^{(i)},\theta^{(j)}$ view pairs.  The outer $N \times M$ grid considers each  $\theta^{(i)}$, and each inner $N \times M$ grid considers each $\theta^{(j)}$ for the given $\theta^{(i)}$ (bottom left).  A pixel in that grid is bright if coverage is high for $\theta^{(j)}$ given $\theta^{(i)}$, and dark otherwise.  Each $\theta^{(i)}$ denotes an (elevation, azimuth) pair.  
While observed views and their neighbors are naturally recoverable (brighter), the sidekick uses broader environment context to also anticipate distant and/or different-looking parts of the environment, as seen by the non-uniform spread of scores in the left grid.
    Given the coverage function and a starting position, this sidekick selects actions to greedily optimize the coverage objective (Eq~\ref{coverage_objective}).
    The bottom right strip shows the cumulative coverage maps as each of the $T$=4 glimpses is selected.}
    \label{sidekick-framework}
\end{figure}

A candidate view is scored based on how informative it is, i.e., how well the
entire environment can be reconstructed given only that view. We train a completion model (cf.~Sec.~\ref{sec:completion}) 
that can reconstruct $\hat{V}(X)$ from any single view (i.e., we set $T=1$).  Let $\hat{V}(X | y)$ denote the decoded reconstruction for $X$ given only view $y$ as input.
The sidekick scores the information in observation $x(X, \theta)$ as:
\begin{equation}
  \label{one_view_score}
  \text{Info}\left(x(X, \theta), X\right) ~~\propto^{-1}~~d\left(\hat{V}(X|x(X, \theta)), V(X)\right),
\end{equation}
where $d$ denotes the reconstruction error and $V(X)$ is the fully observed environment.  We use a simple $\ell_2$ loss on pixels for $d$ to quantify information. 
Higher-level losses, e.g., for detected objects, could be employed when available.
The scores are normalized to lie in $[0, 1]$ across the different views of $X$. 
The sidekick scores each candidate view.  
Then, in order to sharpen the effects of the scoring function and avoid favoring redundant observations, the sidekick selects the top $K$ most informative views with greedy non-maximal suppression.  It iteratively selects
the view with the highest score and suppresses all views in the neighborhood of that view until $K$ views are selected (see Supp.~for details).  This yields a map of favored views for each training environment.  See Fig~\ref{sidekick-framework}, top row.

The sidekick conveys the results to the agent during policy learning in the form of an augmented reward (to be defined in Sec.~\ref{sec:obj}).  Thus, the reward-based sidekick previews observations and encourages the selection of those \emph{individually} valuable for reconstruction.
Note that while the sidekick indexes views in absolute angles, the agent will not; all its observations are relative to its initial (random) glimpse direction.  This works because the sidekick becomes a part of the environment, i.e., it attaches rewards to the true views of the environment.  
In short, the reward-based sidekick shapes rewards based on its exploration with full observability.  
\vspace{-0.2cm}

\subsubsection{Demonstration-based sidekick}
\label{sec:demo}
Our second sidekick generates \emph{trajectories} of informative views.  Given a starting
view in $X$, the demonstration sidekick selects a trajectory of $T$ views that are deemed to be most informative about $X$. 
Unlike the reward-based sidekick above, this sidekick offers guidance with respect to a starting state, and it is subject to the same camera motion restrictions placed on the main agent.  Such restrictions model how an agent cannot teleport its camera using one unit of effort.

To identify informative trajectories, we first define a
scoring function that captures \emph{coverage}.  Coverage reflects how much information $x(X, \theta)$ contains about each view in $X$. 
The coverage score for view $\theta^{(j)}$
upon selecting view $\theta^{(i)}$ is:
\begin{equation}
  \label{coverage}
  \text{Coverage}_{X}\left(\theta^{(j)} | \theta^{(i)}\right) \propto^{-1} d\left(\hat{x}(X, \theta^{(j)}), x(X, \theta^{(j)}) \right),
\end{equation}
where $\hat{x}$ denotes an inferred view within $\hat{V}(X | x(X, \theta^{(i)}))$, as estimated using the same $T=1$ completion network used by the reward-based sidekick.
Coverage scores are normalized to lie in $[0,1]$ 
for  $ 1 \le i, j \le MN$.\vspace*{-0.10in}  
\begin{equation}
  \label{coverage_objective}
  \mathcal{C}(\Theta, X) = \sum_{j=1}^{MN} \sum_{\theta\in\Theta} \text{Coverage}_{X}(\theta^{(j)} | \theta),
\end{equation}
The goal of the demonstration sidekick is to maximize the coverage objective (Eqn.~\ref{coverage_objective}), where $\Theta  = \{\theta_{1}, \ldots, \theta_{t}\}$ denotes the sequence of selected views, and $\mathcal{C}(\Theta, X)$ saturates at 1.
In other words, it seeks a sequence of reachable views such that \emph{all} views are ``explained" as well as possible.  See Fig.~\ref{sidekick-framework}, bottom panel. 

The policy of the sidekick ($\pi_{s}$) is to greedily select actions based on the coverage objective.  The objective encourages the sidekick to select views such that the overall
information obtained about each view in $X$ is maximized.  
\begin{equation}
  \label{expert_policy}
  \pi_{s}(\Theta) = \argmax_{\delta}~\mathcal{C}\left(\Theta \cup \{\theta_{t} + \delta\}, X\right).
\end{equation}

We use these sidekick-generated trajectories as supervision to the agent for a short preparatory period. The goal is to initialize
the agent with useful insights learned by the sidekick to accelerate training of better policies.
We achieve this through a hybrid training procedure that combines imitation and reinforcement.
In particular, for the first $t_{sup}$ time steps,
we let the sidekick drive the action selection and train the policy based on a supervised objective. 
For steps $t_{sup}$ to $T$,
we let the agent's policy drive the action selection and use REINFORCE~\cite{williams1992simple} or Actor-Critic~\cite{sutton1998reinforcement} to update the agent's policy (see Sec.~\ref{sec:experiments}).  We start with $t_{sup} = T$ and gradually reduce it to $0$ in the preparatory sidekick phase (see Supp.).
This step relates to behavior cloning~\cite{bojarski2016end,giusti2016machine,duan2017one}, which formulates policy learning as supervised action classification given states.  However, unlike typical behavior cloning, the sidekick is not an expert.  It solves a simpler version of the task, then backs away as the agent takes over to train with partial observability.
\vspace{-0.1cm}

\subsection{Policy learning with sidekicks}
\label{sec:obj}

Having defined the two sidekick variants, we now explain how they influence policy learning.
The goal is to learn the policy $\pi(\delta | a_t)$ which returns a distribution over actions for the aggregated internal representation $a_t$ at time $t$.
Let $\mathcal{A} = \{\delta_i\}$ denote the set of camera motions available to the agent.

Our agent seeks the policy that minimizes reconstruction error for the environment given a budget of $T$ camera motions (views). If we denote the set of 
weights of the network $[W_{s}, W_{f}, W_{r}, W_{d}, W_{a}]$ by $W$ and $W$ excluding $W_{a}$ by $W_{/a}$ and $W$ exluding $W_{d}$ by $W_{/d}$, then 
the overall weight update is:

\begin{equation}
\Delta W = \frac{1}{n} \sum_{j=1}^{n} \lambda_{r} \Delta W^{rec}_{/a} + \lambda_{p} \Delta W^{pol}_{/d}
\end{equation}

where $n$ is the number of training samples, $j$ indexes over the training samples, $\lambda_r$ and $\lambda_p$ are constants and $\Delta W^{rec}_{/a}$ and $\Delta W^{pol}_{/d}$ update all parameters except $W_{a}$ and $W_{d}$, respectively. The pixel-wise MSE reconstruction loss ($\mathcal{L}^{rec}_{t}$) and corresponding weight update at time $t$ are given in Eqn.~\ref{rec_loss}, where $\hat{x}_{t}(X, \theta^{(i)})$ denotes the reconstructed view at viewpoint $\theta^{(i)}$ and time $t$, and $\Delta_{0}$ denotes the offset to account for the unknown starting azimuth (see~\cite{dinesh-ltla}).
\begin{equation}
\label{rec_loss}
\begin{split}
  \mathcal{L}_{rec}^t(X) =  \sum_{i=1}^{MN} d\left(\hat{x}_{t}(X, \theta^{(i)}+\Delta_{0}), x(X, \theta^{(i)})\right), \\
  \Delta W^{rec}_{/a} = -\sum_{t=1}^{T} \nabla_{W_{/a}} \mathcal{L}_{rec}^{t}(X),
\end{split}
\end{equation}

The agent's reward at time $t$ (see Eqn.~\ref{rec_reward}) consists of the intrinsic reward from the sidekick $r^{s}_t = \text{Info}(x(X,\theta_t),X)$ (see Sec.~\ref{sec:reward}) and the negated final reconstruction loss ($-\mathcal{L}_{rec}^T(X)$).
\begin{equation}
\label{rec_reward}
r_{t} = \begin{cases}
          r^{s}_{t} &\quad 1 \le t \le T-2\\
         
          -\mathcal{L}_{rec}^T(X) + r^{s}_{t} &\quad t = T-1\\
        \end{cases}
\end{equation}
The update from the policy (see Eqn.~\ref{eqn:reinforce}) consists of the REINFORCE update, with a baseline $b$ to reduce variance, and supervision from the demonstration sidekick (see Eqn.~\ref{supervised_loss}). We consider both REINFORCE~\cite{williams1992simple} and Actor-Critic~\cite{sutton1998reinforcement} methods to update the \textsc{Act} module.  For the latter, the policy term additionally includes a loss to update a learned Value Network (see Supp.).  For both, we include a standard entropy term to promote diversity in action selection and avoid converging too quickly to a suboptimal policy.
\begin{equation}
\label{eqn:reinforce}
\Delta W_{/d}^{pol} = \sum_{t=1}^{T-1} \nabla_{W_{/d}} \text{log}\,\pi(\delta_{t}|a_{t})\bigg(\sum_{t^{'}=t}^{T-1}r_{t^{'}} - b(a_{t})\bigg) + \Delta W_{/d}^{demo},\vspace{-0.10cm}
\end{equation}

The demonstration sidekick influences policy learning via a cross entropy loss between the sidekick's policy $\pi_s$ (cf.~Sec.~\ref{sec:demo}) and the agent's policy $\pi$:\vspace{-0.10cm}
\begin{equation}
\label{supervised_loss}
\Delta W_{/d}^{demo} = \sum_{t=1}^{T-1} \sum_{\delta \in \mathcal{A}} \nabla_{/d}(\pi_s(\delta | a_{t})~ \text{log}\,\pi(\delta | a_{t})).\vspace{-0.10cm}
\end{equation}

We pretrain the \textsc{Sense}, \textsc{Fuse}, and \textsc{Decode} modules
with $T=1$. The full network is then trained end-to-end (with \textsc{Sense} and \textsc{Fuse} frozen). For training with sidekicks, the agent is augmented either with additional rewards from the reward sidekick (Eqn.~\ref{rec_reward}) or an additional supervised loss from the demonstration sidekick (Eqn.~\ref{supervised_loss}). As we will show empirically, training with sidekicks helps overcome uncertainty due to partial observability and learn better policies. 
\vspace{-0.1cm}

\subsection{Visualizing the learned motion policies}
\label{sec:vis}

Finally, we propose a visualization technique to qualitatively understand the policy that has been learned. The aggregated state $a_{t}$ is
used by the policy network to determine the action probabilities. To analyze which part of the agent's belief ($a_{t}$) is important
for the current selected action $\delta_{t}$, we solve for the change in the aggregated state ($\Delta a_{t}$) which maximizes the
change in the predicted action distribution ($\pi(\cdot | a_{t})$):\vspace{-0.10cm}
\begin{equation}
  \centering
  \label{eqn_vis}
  \begin{split}
        \Delta a^{*} = \argmax_{\Delta a_{t}} \sum_{\delta \in \mathcal{A}} \big( \pi(\delta | a_t) - \pi(\delta | a_t + \Delta a_t)\big)^2\\
        s.t.~||\Delta a_{t}|| \le C||a_{t}||
  \end{split}
\end{equation}
where $C$ is a constant that limits the deviation in norm from the true belief. Eqn.~\ref{eqn_vis} is maximized using gradient ascent (see Supp.). This change in belief is visualized in the viewgrid space by forward propagating through the \textsc{Decode} module. The visualized heatmap intensities ($H_{t}$) are defined as follows:
\begin{equation}
  H_{t} \propto ||\textsc{Decode}(a_{t} + \Delta a^{*}) - \textsc{Decode}(a_{t})||^{2}_{2}.
\end{equation}
The heatmap indicates which parts of the agent's belief \emph{would have to change} to affect its action selection. The views with high intensity are those that affect the agent's action selection the most.

\section{Experiments}
\label{sec:experiments}
In Sec.~\ref{sec:dataset},~\ref{sec:results}, we describe our experimental setup and analyze the learning efficiency and test-time performance of different methods. In Sec.~\ref{sec:vis_exp}, we visualize learned policies and demonstrate the superiority of our policies over a baseline.
\vspace{-1em}
\subsection{Experimental Setup}
\label{sec:dataset}

\noindent\textbf{Datasets:} We use two popular datasets to benchmark our models. 

\begin{itemize}
\item \textbf{SUN360:} SUN360~\cite{xiao2012recognizing} consists of high resolution spherical panoramas from multiple scene 
categories. We restrict our experiments to the 26 category subset used in~\cite{xiao2012recognizing,dinesh-ltla}.
The viewgrid consists of 32$\times$32 views captured across 4 elevations (-$45^{\circ}$ to $45^{\circ}$) and 8 azimuths ($0^{\circ}$ to $180^{\circ}$).
At each step, the agent sees a  
$60^{\circ}$ field-of-view.  
This dataset represents an agent looking out at a scene in a series of narrow field-of-view glimpses. \vspace{0.5em}
\vspace{-0.3em}
\item\textbf{ModelNet Hard:} ModelNet~\cite{shapenet} provides a collection of 3D CAD models for different categories of
objects. ModelNet-40 and ModelNet-10 are provided subsets consisting of 40 and 10 object categories 
respectively, the latter being a subset of the former. We train on objects from the 30 categories not present in ModelNet-10 and test on objects from the unseen 10 categories.
We increase completion difficulty in ``ModelNet Hard'' by rendering with more challenging lighting conditions, textures and viewing angles than~\cite{dinesh-ltla}; see Supp. It consists of $32\times 32$ views sampled from 5 elevations and 9 azimuths.  This dataset represents an agent looking in at a 3D object and moving it to a series of selected poses.
\end{itemize}
\vspace{-0.15em}
\noindent For both datasets, the candidate motions $\mathcal{A}$ are restricted to a 3 elevations x 5 azimuths neighborhood, representing the set of unit-cost actions. Neighborhood actions mimic real-world scenarios where the agent's physical motions are constrained (i.e., no teleporting) and is consistent with recent active vision work~\cite{dinesh-ltla,germs-bmvc2015,dinesh-eccv2016,dinesh-pami2018,ammirato-icra2017}. The budget for number of steps is fixed to $T=4$.\\

\vspace{-0.25em}
\noindent\textbf{Baselines:} We benchmark our methods against several baselines:
\begin{itemize}
\item\texttt{one-view}: the agent trained to reconstruct from one view ($T=1$).
\item\texttt{rnd-actions}: samples actions uniformly at random.
\item\texttt{ltla}~\cite{dinesh-ltla}: our implementation of the ``learning to look around'' approach~\cite{dinesh-ltla}.  We verified our code reproduces results from~\cite{dinesh-ltla}.
\item\texttt{rnd-rewards}: naive sidekick where rewards are assigned uniformly at random on the viewgrid.
\item\texttt{asymm-ac}~\cite{pinto2017asymmetric}: approach from~\cite{pinto2017asymmetric} adapted for discrete actions. Critic sees the entire panorama/object and true camera poses (no experience replay).
\item\texttt{demo-actions}: actions selected by demo-sidekick while training / testing. 
\item\texttt{expert-clone}: imitation from an expert policy that uses full observability (similar to critic in Fig.~2 of Supp.) 
\end{itemize}
\vspace{-0.2em}

\noindent\textbf{Evaluation:} We evaluate reconstruction error averaged over uniformly sampled elevations, azimuths and all test samples (\texttt{avg}). To provide a worst case analysis,
we also report an adversarial metric (\texttt{adv}), which evaluates each agent on its hardest starting positions in each test sample and averages over the test data.

\subsection{Active Exploration Results} 
\label{sec:results}

\begin{table}[t]
\centering
\scalebox{0.85}{
\begin{tabular}{| p{3.5cm} || P{3.5em} P{3.5em} | P{3.5em} P{3.5em}|| P{3.5em} P{3.5em} | P{3.5em} P{3.5em} |}
\hline
\multirow{2}{*}{Method}                                & \multicolumn{4}{c||}{SUN360}                                             & \multicolumn{4}{c|}{ModelNet Hard}                                          \\ \cline{2-9} 
                                                       & \multicolumn{2}{c|}{avg ($\times$1000)}   & \multicolumn{2}{c||}{adv ($\times$1000)}   & \multicolumn{2}{c|}{avg ($\times$1000)}  & \multicolumn{2}{c|}{adv ($\times$1000)}   \\
                                                       & mean $\downarrow$  & $\%\uparrow$                      & mean $\downarrow$ & $\%\uparrow$                      & mean $\downarrow$ & $\%\uparrow$                      & mean $\downarrow$ & $\%\uparrow$                      \\ \hline
\texttt{one-view}                                      & 38.31 & -                                 & 55.12 & -                                 & 9.63 & -                                 & 17.10 & -                                 \\
\texttt{rnd-actions}                                   & 30.99 & 19.09                             & 44.85 & 18.63                             & 7.32 & 23.93                             & 12.38 & 27.56                             \\
\texttt{rnd-rewards}                                   & 25.55 & 33.30                             & 30.20 & 45.21                             & 7.04 & 26.89                             & 9.66  & 43.50                             \\
\texttt{ltla~\cite{dinesh-ltla}} & 24.94 & 34.89                             & 31.86 & 42.19                             & 6.30 & 34.57                             & 8.78  & 48.65                             \\
\texttt{asymm-ac~\cite{pinto2017asymmetric}}                                      & 23.74 & 38.01                             & 29.92 & 45.72                             & 6.24 & 35.20                             & 8.55  & 50.00                             \\
\texttt{expert-clone}             & 23.98 & 37.38         & 28.50 & \textcolor{Blue}{\textbf{48.28}}         & 6.41  & 33.44         & 8.52 & 50.13         \\ 
\hline
\texttt{ours(rew)}                                     & 23.44 & \textcolor{Blue}{\textbf{38.82}}                             & 28.54 & 48.22                             & 5.80 & \textcolor{Blue}{\textbf{39.79}}  & 7.17  & \textcolor{Blue}{\textbf{58.04}} \\
\texttt{ours(demo)}                                    & 24.24 & 36.73                             & 29.01 & 47.36                             & 6.32 & 34.37                             & 8.64  & 49.47                             \\
\texttt{ours(rew)+ac}                                & 23.36 & \textcolor{Green}{\textbf{39.01}} & 28.26 & \textcolor{Green}{\textbf{48.72}}  & 5.75 & \textcolor{Green}{\textbf{40.26}} & 7.10  & \textcolor{Green}{\textbf{58.44}} \\
\texttt{ours(demo)+ac}                               & 24.05 & 37.22  & 28.52 & 48.26 & 6.13 & 36.31                             & 8.26  & 51.64                             \\
\hline
\rowcolor[gray]{0.9} \texttt{demo-actions}\textsuperscript{*}                                      & 26.12 & 31.82                             & 31.53 & 42.76                             & 5.82 & 39.50                             & 7.46  & 56.40                             \\
\hline
\end{tabular}
}
\caption{Avg/Adv MSE errors $\times 1000$ ($\downarrow$ lower is better) and corresponding improvements (\%) over the \texttt{one-view} model ($\uparrow$ higher is better), for the two datasets.  
 The best and second best performing models are highlighted in \textcolor{Green}{green} and \textcolor{Blue}{blue} respectively. Standard errors range from 0.2 to 0.3 on SUN360 and 0.1 to 0.2 on ModelNet Hard. (* - requires full observability at test time)}
\label{tab:quant}
\vspace{-0.5cm}
\end{table}

Table~\ref{tab:quant} shows the results on both datasets.  For each metric, we report the mean error along with the percentage improvement over the \texttt{one-view} baseline.  
Our methods are abbreviated \texttt{ours(rew)} and \texttt{ours(demo)} referring to the use of our reward- and demonstration-based sidekicks, respectively.  We denote the use of Actor-Critic instead of REINFORCE with \texttt{+ac}.

\begin{figure}[t!]
    \centering
    \scalebox{1}{
    \includegraphics[width=0.90\textwidth, clip, trim={0.5cm 9.8cm 4cm 0.2cm}]{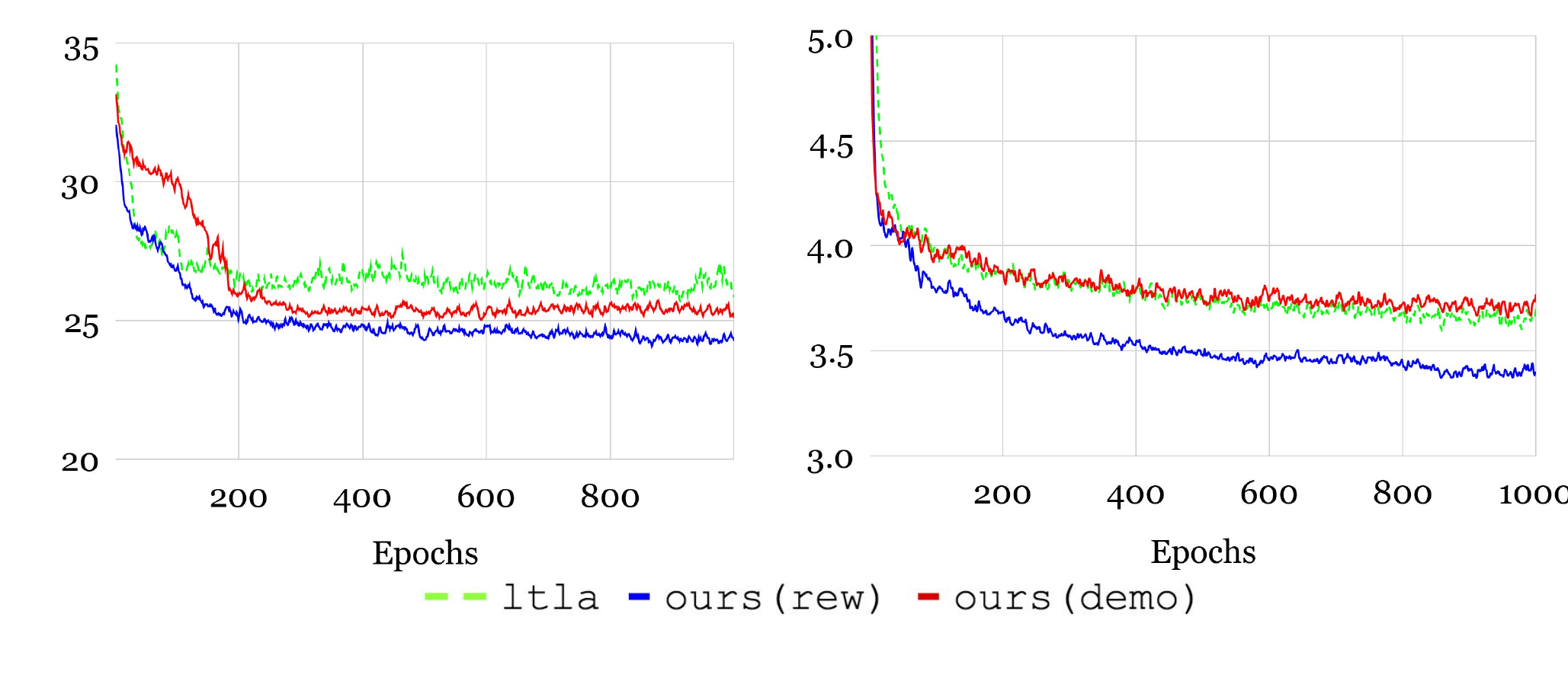}
    }
    \vspace*{-0.15in}
    \caption{\footnotesize Validation errors ($\times 1000$) vs.~epochs on SUN360 (left) and ModelNet Hard (right).  All models shown here use REINFORCE (see Supp.~for more curves).  Our approach accelerates convergence.  }\vspace*{-0.15in}
    \label{val_plots}
    
\end{figure}

We observe that \texttt{ours(rew)} and \texttt{ours(demo)} with REINFORCE generally perform better than \texttt{ltla} with REINFORCE~\cite{dinesh-ltla}. In particular, \texttt{ours(rew)} performs significantly better 
than \texttt{ltla} on both datasets on all metrics. \texttt{ours(demo)} performs better on SUN360, but is only slightly better on ModelNet Hard.
Figure~\ref{val_plots} shows the validation loss plots; using the sidekicks leads to significant improvement in the convergence rate over \texttt{ltla}.

\begin{figure}[t!]
    \centering
      \includegraphics[width=0.99\textwidth, clip, trim={0.5cm 0cm 2cm 0.5cm}]{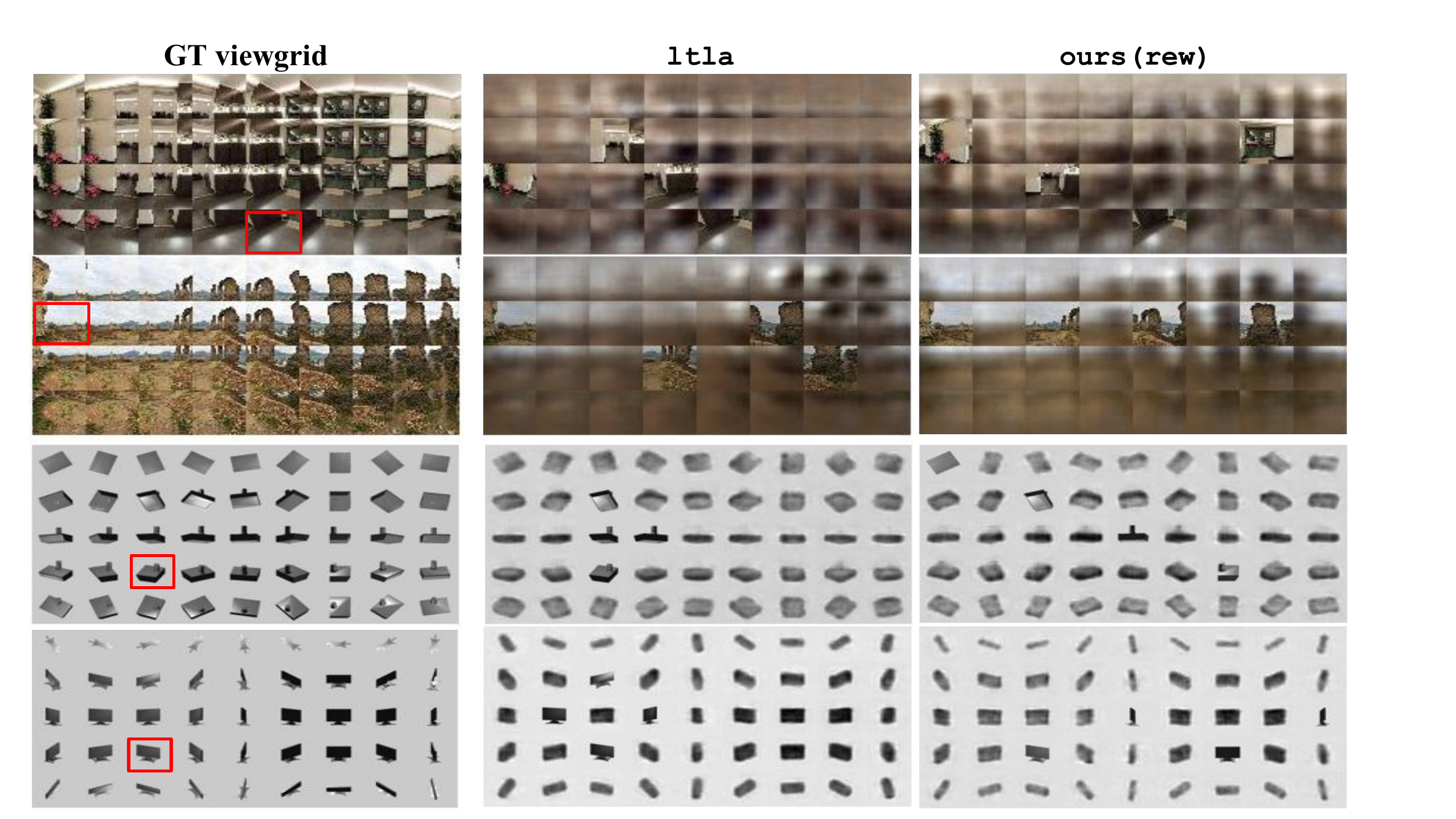}
    \caption{\footnotesize Qualitative comparison of \texttt{ours(rew)} vs.~\texttt{ltla}~\cite{dinesh-ltla} on SUN360 (first 2 rows) and
    ModelNet Hard (last 2 rows). The first column shows the groundtruth viewgrid and a randomly selected starting point (marked in red). 
    The 2nd and 3rd columns contain the decoded viewgrids from \texttt{ltla} and \texttt{ours(rew)} after $T=4$ time steps. The reconstructions
   from \texttt{ours(rew)} are visibly better.  
   For example, in the $3^{rd}$ row, our model reconstructs the protrusion more clearly; in the $2^{nd}$
    row, our model reconstructs the sky and central hills more effectively.  Best viewed on pdf with zoom.
    }
    \label{sun360_qual}
    \vspace{-0.2in}
\end{figure}

Figure~\ref{sun360_qual} compares example decoded reconstructions.  We stress that the vast majority of pixels are unobserved when decoding the belief state, i.e., only $4$ views out of the entire
viewing sphere are observed. Accordingly, they are blurry.  Regardless, their differences indicate the differences in belief states between the two methods.  A better policy more quickly fleshes out the general shape of the scene or object.

Next, we compare our model to \texttt{asymm-ac}, which is an alternate paradigm for exploiting full observability during training. First, we note that \texttt{asymm-ac} performs better than \texttt{ltla} across all datasets and metrics
, making it a strong baseline.
Comparing \texttt{asymm-ac} with \texttt{ours(rew)+ac} and \texttt{ours(demo)+ac}, we see our methods still perform considerably better on all metrics and datasets.  
As we show in the Supp, our methods also lead to faster convergence.

In order to contrast learning from sidekicks with learning from experts, we additionally compare our models to behavior cloning an expert that exploits full observability at training time. As shown in Tab.~\ref{tab:quant}, \texttt{ours(rew)} outperforms \texttt{expert-clone} on both the datasets, validating the strength of our approach. It is particularly interesting because training an expert takes a lot longer ($17\times$) than training sidekicks (see Supp.). When compared with \texttt{demo-actions}, an ablated version of \texttt{ours(demo)} that requires full observability at \emph{test time}, our performance is still significantly better on SUN360 and slightly better on ModelNet Hard.  \texttt{ours(rew)} and \texttt{ours(demo)} also beat the remaining baselines by a significant margin. These results verify our hypothesis that sidekick policy learning can improve over strong baselines by exploiting full observability during training.

\vspace{-0.1in}
\subsection{Policy Visualization}
\label{sec:vis_exp}
\begin{figure}[t]
    \centering
      \includegraphics[width=\textwidth, clip, trim={0cm 3.5cm 1cm 0cm}]{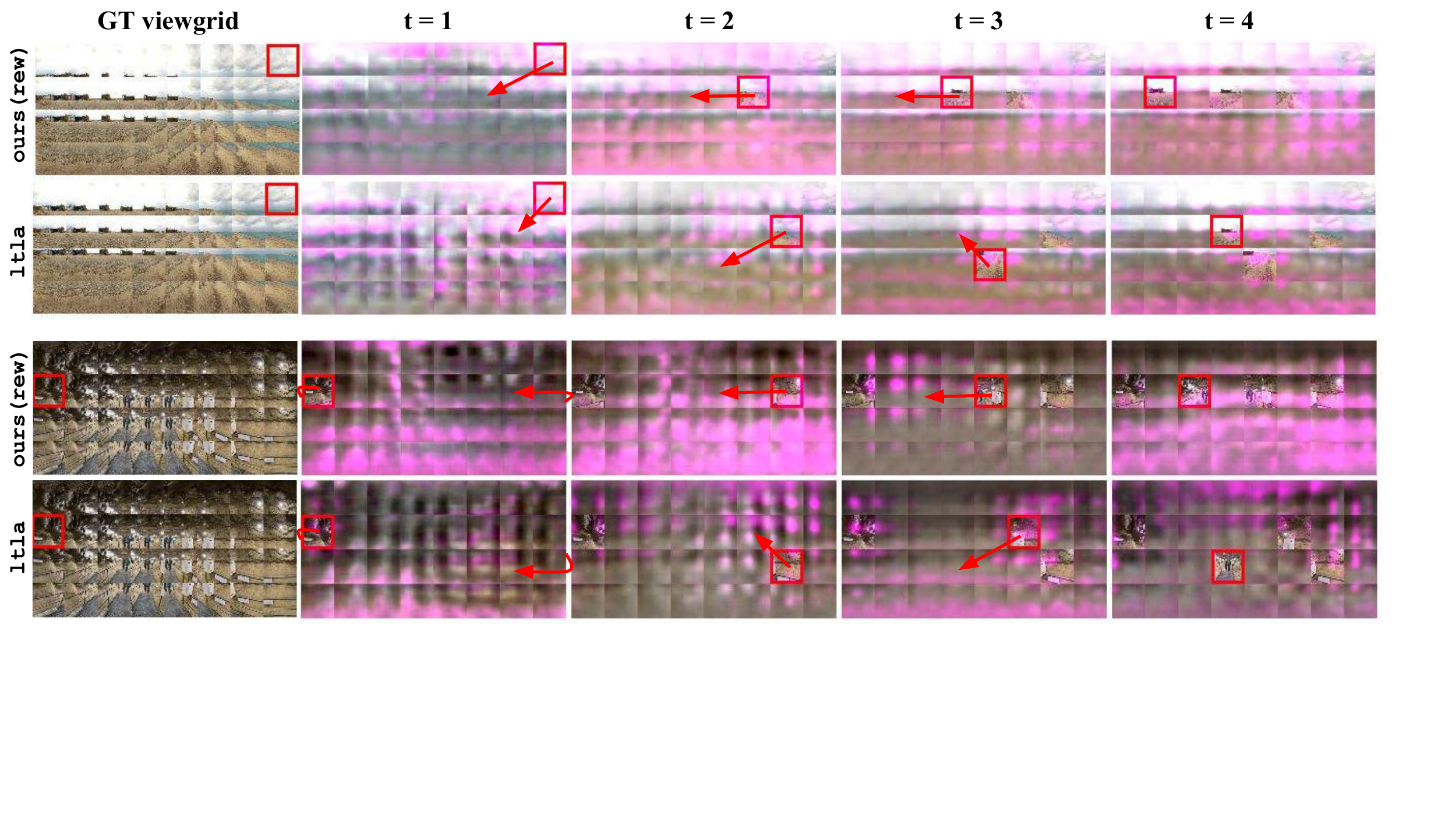}
    \caption{\footnotesize Policy visualization: The viewgrid reconstructions of \texttt{ours(rew)} and \texttt{ltla}~\cite{dinesh-ltla} are shown on 
    two examples from SUN360. The first column shows the viewgrid with a randomly selected view (in red).  Subsequent columns show the
    view received (in red), viewgrid reconstructed, action selected (red arrow), and the parts of the belief space our method deems responsible for the action selection (pink heatmap). 
    Both the agents tend to move towards sparser regions of the heatmap, attempting to improve their beliefs about views that do not contribute to their action selection. \texttt{ours(rew)} improves its beliefs much more rapidly and as a result, performs more informed action selection.}
    \label{sun360_vis}
\end{figure}

\vspace{-0.1in}
We present our policy visualizations for \texttt{ltla} and \texttt{ours(rew)} on SUN360 in Figure~\ref{sun360_vis}; see Supp.~for examples with \texttt{ours(demo)}.
The heatmap from Eq~\ref{eqn_vis} is shown in pink and overlayed on the reconstructed viewgrids.
For both models, the policies tend to take actions that move them towards views which have low heatmap density, as witnessed
by the arrows / actions pointing to lower density regions. 
Intuitively, the agents move towards the views that are not contributing effectively to their action selection to
increase their understanding of the scene. It can observed in many cases that \texttt{ours(rew)} model has a much denser
heat map across time when compared to \texttt{ltla}. Therefore, \texttt{ours(rew)} takes more views into account for
selecting its actions earlier in the trajectory, suggesting that a better policy and history aggregation leads to more informed action selection. 

\section{Conclusion}

We propose \textit{sidekick policy learning}, a framework to leverage extra observability or fewer restrictions on an agent's motion during training to learn better policies.
We demonstrate the superiority of policies 
learned with sidekicks on two challenging datasets, improving over existing methods and accelerating training. Further, we 
utilize a novel policy visualization technique to illuminate the different reasoning behind policies trained with and without sidekicks.  In future work, we plan to investigate the effectiveness of our framework on other active vision tasks such as recognition and navigation. 

\section*{Acknowledgements}

The authors thank Dinesh Jayaraman, Thomas Crosley, Yu-Chuan Su, and Ishan Durugkar for helpful discussions.
This research is supported in part by DARPA Lifelong Learning Machines, a Sony Research Award, and an IBM Open Collaborative Research Award. 
%
%
%
\bibliographystyle{splncs04}
\bibliography{egbib}
\title{Sidekick Policy Learning\\for Active Visual Exploration \\ Supplementary Material}

\titlerunning{Sidekick Policy Learning for Active Visual Exploration}

\authorrunning{S. Ramakrishnan and K. Grauman}

\author{Santhosh K. Ramakrishnan\inst{1} \and Kristen Grauman\inst{2}}
\institute{The University of Texas at Austin, Austin, TX 78712 \and
           Facebook AI Research, 300 W. Sixth St. Austin, TX 78701 \\
           \email{srama@cs.utexas.edu, grauman@fb.com\thanks{\it{On leave from University of Texas at Austin (grauman@cs.utexas.edu)}}} \\ 
           Project page: \url{ http://vision.cs.utexas.edu/projects/sidekicks/}}

\maketitle

In this document, we provide the following additional information:
\begin{enumerate}
  \item Architectures and implementation details.
  \item Additional policy learning details.
  \item Validation plots.
  \item ModelNet Hard dataset construction.
  \item Additional policy visualization examples.
  \item Additional training time for sidekicks.
\end{enumerate}


\section{Architectures and implementation details}
\label{sec:arch}
\begin{figure}[h]
  \includegraphics[width=\textwidth]{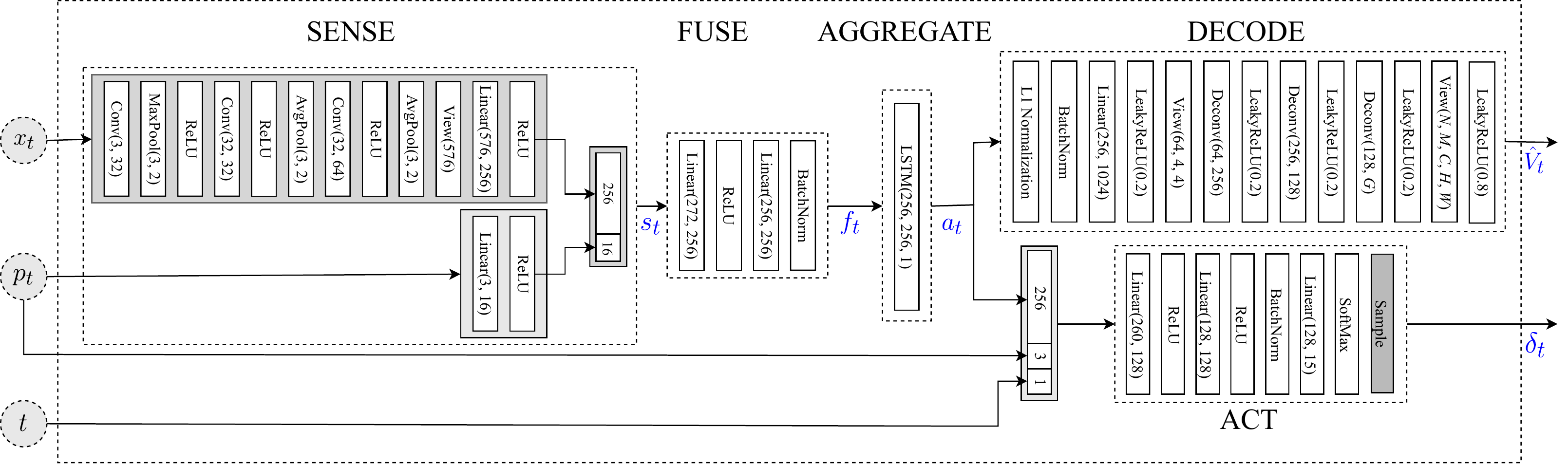}
  \caption{Architecture for \texttt{ltla} baseline~\cite{dinesh-ltla}. Note: G = M*N*C}
  \label{fig:arch_ltla}
\end{figure}
\begin{figure}[h]
  \includegraphics[width=\textwidth]{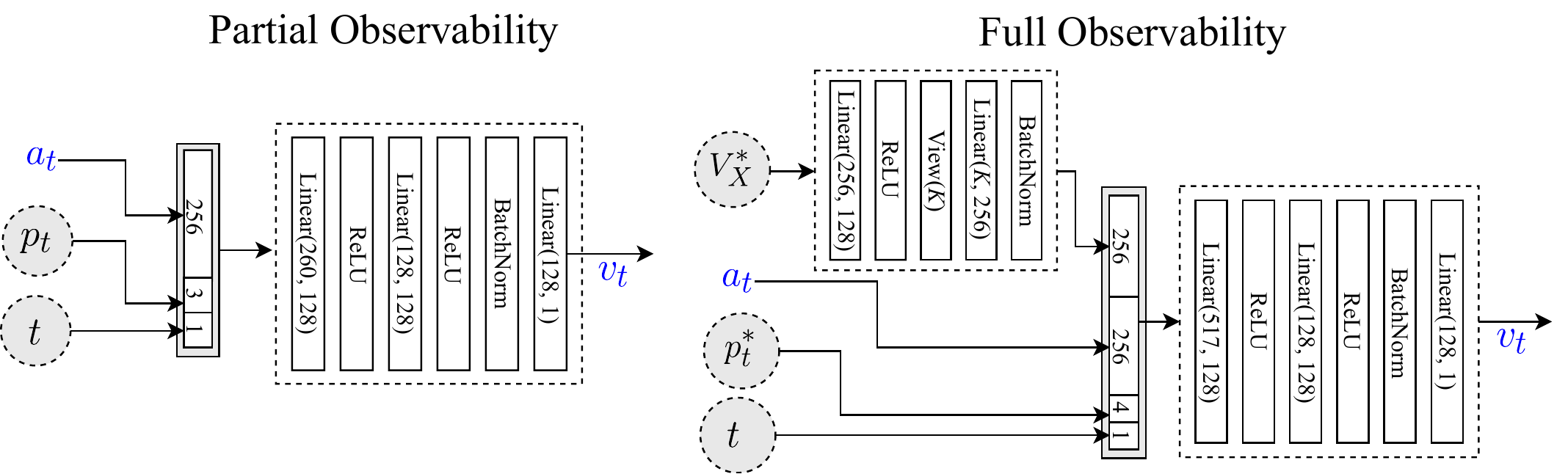}
  \caption{Architecture for critics used in our Actor Critic training. The Partial Observability critic is used for \texttt{ours(rew)+ac}, \texttt{ours(demo)+ac} and the Full Observability critic is used for \texttt{asymm-ac}~\cite{pinto2017asymmetric}. Note: $K = N*M*128$}
  \label{fig:arch_critic}
\end{figure}
Before we review the architecture, we list out some key notations:
\begin{itemize}
  \item{$p_{t}$ - proprioception input, consists of the relative change in elevation, azimuth from $t-1$ to $t$ and the absolute elevation at $t$. }
  \item{$p_{t}^{*}$ - $p_{t}$ augmented with absolute azimuth}
  \item{$x_{t}$ - input view, dimensionality is $C\times H \times W$ where $C$ is the number of channels, $H$ is the image height and $W$ is the image width. For SUN360, $C=3, H=32, W=32$ and for ModelNet Hard, $C=1, H=32, W=32$.} 
  \item{$M$ - number of azimuths in $X$ ($8$ for SUN360 and $9$ for ModelNet Hard).}
  \item{$N$ - number of elevations in $X$ ($4$ for SUN360 and $5$ for ModelNet Hard).}
\end{itemize}
We follow the same architecture (see Fig.~\ref{fig:arch_ltla}) for the modules described in~\cite{dinesh-ltla}.
Models are implemented in PyTorch and layer naming conventions are accordingly followed~\footnote{refer to http://pytorch.org/docs/master/nn.html}. 
For all the Conv layers, filter size = 5, stride = 1 and zero padding = 2; for all the 
Deconv (aka transposed convolution) layers, filter size = 5, stride = 2, zero padding = 2 and output 
padding = 1.

We have two critic architectures for our experiments (see Fig.~\ref{fig:arch_critic}). The critic with partial observability consists of a similar architecture as the \textsc{Act} module. The critic with full observability takes in the absolute position on the viewgrid and the entire viewgrid as additional
inputs. Each view of the viewgrid is processed by the \textsc{Sense} module (to give $V_{X}^{*}$) and the encoded views are fused together using two FC layers. This aggregated
state, proprioception input, absolute position, and fused viewgrid are concatenated and processed by the critic to obtain the value of the current view. 

We use the Adam optimizer with a learning rate of $0.0001 - 0.003$, weight decay of 1e-6,
and other default settings from PyTorch~\footnote{refer http://pytorch.org/docs/master/optim.html}. We also set $\lambda_{r}=1$ and $\lambda_{p}=1$ based on grid search. In the case of the demonstration-based sidekick, we decay $T_{sup}$ from $T-1$ to $0$ after every 50 epochs. For the reward-based sidekick, we decay the rewards by a factor of $2-10$ after every $100-500$ epochs (selected based on grid search). All the models are trained for $1000$ epochs. For the reward-based sidekick, we use a non-maximal suppression neighborhood of $1$ and $K = 4$ views for SUN360, and neighborhood of $2$ and $K = 5$ views for ModelNet Hard. The neighborhood and number of views were selected manually upon brief visual inspection to ensure sufficient spread of rewards on the viewgrid. 

To solve for $\Delta a^{*}$ from Eq. $10$ in the main paper, we use stochastic gradient descent with learning rate of $0.0001$, weight decay of $0.1$ and momentum of $0.9$. We run the optimization for a maximum of 200 iterations, and perform early stopping if $|\Delta a|$ crosses $0.75|a_{t}|$. The parameters were selected to increase the chances of the probability change being maximised.  

\section{Additional policy learning details}
\label{sec:eqns}
Let the weights of the critic be denoted by $W_{c}$. Following standard actor-critic training, a regression loss over the critic's value prediction is additionally used to update the agent's parameters, specifically, $W_{s}, W_{f}, W_{r}, W_{c}$: 
\begin{equation}
\label{eqn:critic_update}
\Delta W_{\{s, f, r, c\}} = -\nabla_{\{s, f, r, c\}} \frac{1}{n}\sum_{i=1}^{n}\sum_{t=1}^{T-1}\bigg(v_{t}^{i} - \sum_{t^{'}=t}^{T-1} r_{t^{'}}^{i}\bigg)^{2},
\end{equation}
where ${n}$ is the number of data samples and $v_{t}^{i}$ is the value estimated by the Value network at time $t$ for the $i^{th}$ data sample. 
We additionally include a standard entropy term to promote diversity in action selection and avoid converging too quickly to a suboptimal policy. The loss term and the corresponding weight update (on $W_{a}, W_{r}, W_{f}, W_{s}$) are as follows:
\begin{equation}
\label{eqn:entropy_term}
\begin{split}
L_{ent} = \frac{1}{n}\sum_{i=1}^{n}\sum_{t=1}^{T-1}\bigg(\sum_{\delta \epsilon \mathcal{A}}\pi(a_{t}^{i}, j)\text{log}~\pi(a_{t}^{i}, \delta)\bigg) \\
\Delta W_{\{a, r, f, s\}} = -\nabla_{\{a, r, f, s}\} L_{ent}.
\end{split}
\end{equation}

\section{Validation plots}
\label{sec:validation}
\begin{figure}[t!]
    \includegraphics[width=\textwidth, clip, trim={0.5cm 8.5cm 2cm 0.2cm}]{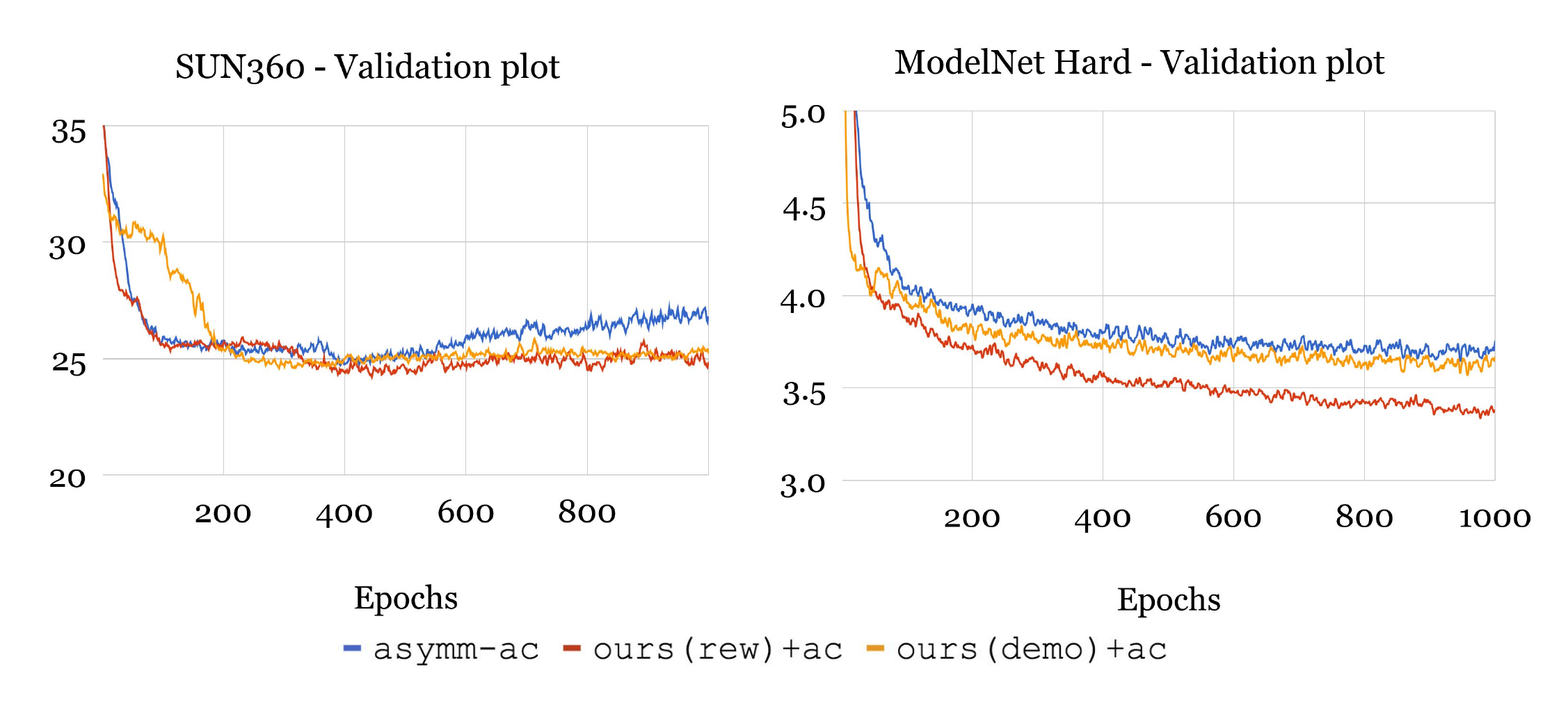}
    \caption{Validation errors ($\times 1000$) vs. epochs on SUN360 and ModelNet Hard (Actor Critic methods)}
    \label{val_plots_2}
\end{figure}

Fig.~4 in the main paper shows the validation error plots for both datasets to compare the speed of learning for our method trained with REINFORCE vs.~\texttt{ltla}~\cite{dinesh-ltla} trained with REINFORCE.
Here, Fig.~\ref{val_plots_2} shows the parallel validation error plots comparing \texttt{ours(rew)}, \texttt{ours(demo)} and \texttt{asymm-ac}~\cite{pinto2017asymmetric} using Actor-Critic.
Note that separating by REINFORCE vs.~Actor-Critic ensures both sets of plots are apples-to-apples.
The bump in the yellow curve on the SUN360 plot reflects how the demonstration schedule changes over epochs. 

\section{ModelNet Hard construction}
\label{sec:modelnet_hard}
\begin{figure}[t!]
    \includegraphics[width=\textwidth, clip, trim={0.5cm 5.5cm 2cm 0.2cm}]{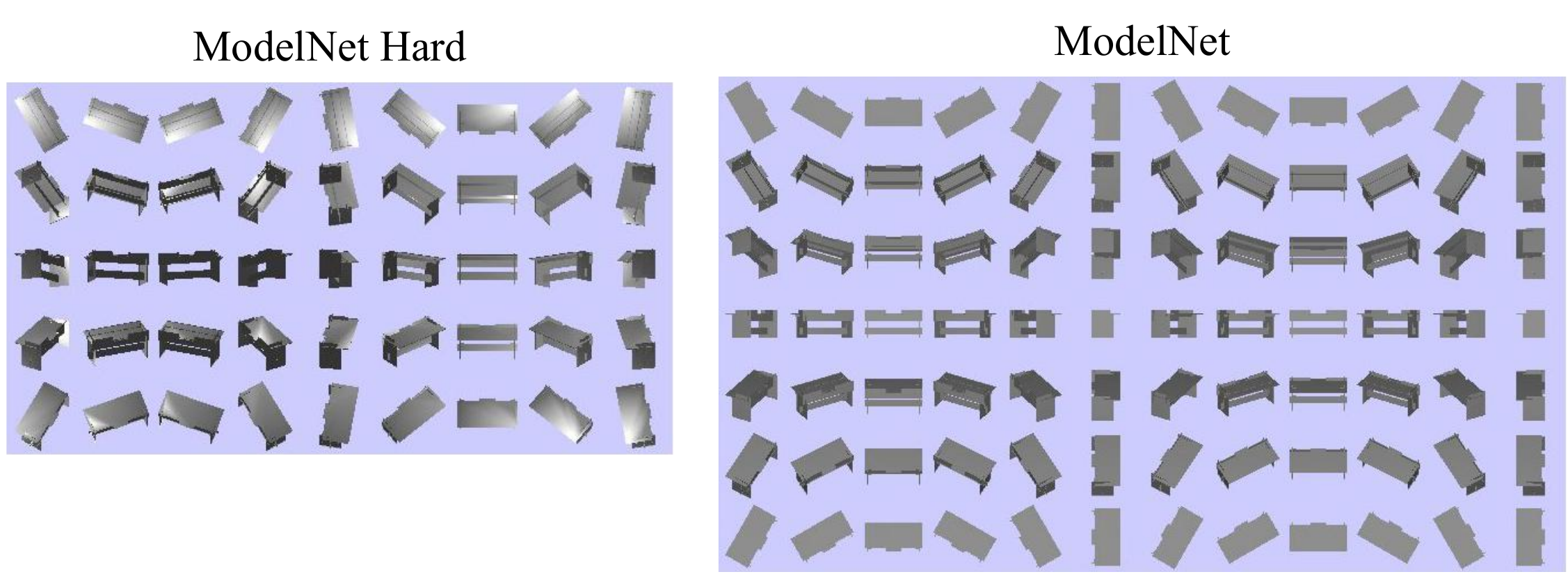}
    \caption{An example to qualitatively compare the renderings of ModelNet Hard vs. ModelNet from~\cite{dinesh-ltla}}
    \label{fig:modelnet_render}
\end{figure}

As noted in the paper, we altered the sampling angles, lighting conditions, and object materials to increase the 
reconstruction difficulty of the rendered images. In Fig.~\ref{fig:modelnet_render}, we render the same object using settings similar to~\cite{dinesh-ltla} and our settings from ModelNet Hard.

The rendering details are as follows.  We sampled the angles at intervals of $40^{\circ}$ (as opposed to $30^{\circ}$ in~\cite{dinesh-ltla}) to reduce the number of views which were similar in appearance and geometry. We further altered the lighting positions to be non-uniform across views and used higher specularity to generate complex renderings. Specifically, we use two light sources, each placed below and above the object. The exact coordinates are selected relative to the size of the object. Each light source is placed randomly at one out of two locations for a given object. Using MATLAB's rendering toolbox\footnote{refer to https://www.mathworks.com/help/matlab/visualize/lighting-overview.html}, we render the objects with ``interp" shading, ``dull" material, ``gouraud" face lighting, ambient strength of 0.4, diffuse strength of 0.9, specular strength of 0.7 and specular exponent of 15. The data is available to ensure reproducibility~\footnote{\url{http://vision.cs.utexas.edu/projects/sidekicks/}}. 


\section{Policy visualization examples}
\label{sec:policy_vis}

\begin{figure}
    \centering
    \begin{subfigure}[b]{0.8\textwidth}
        \includegraphics[width=\textwidth, trim={0cm 4cm 3.5cm 0cm}, clip]{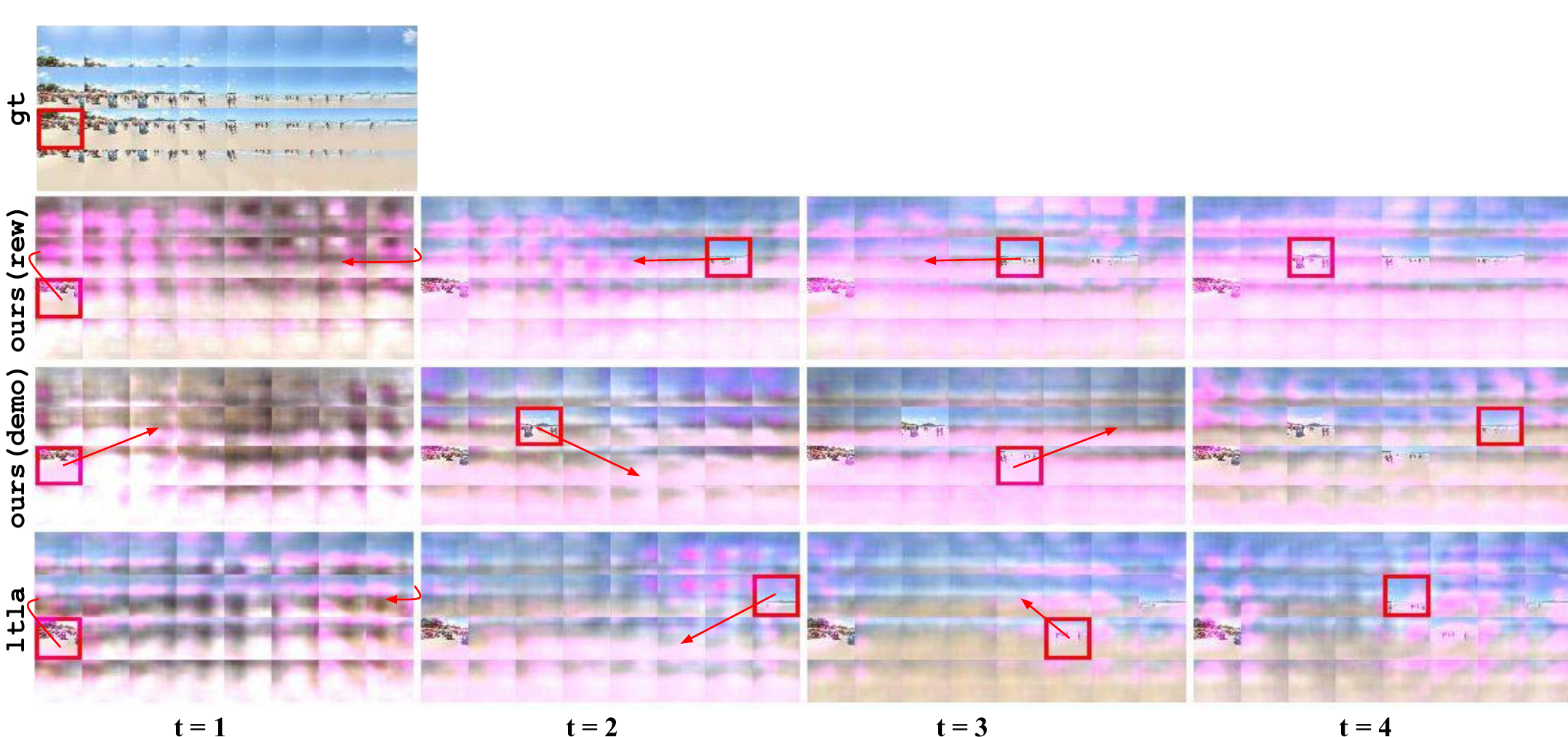}
    \end{subfigure}
    \begin{subfigure}[b]{0.8\textwidth}
        \includegraphics[width=\textwidth, trim={0cm 4cm 3.5cm 0cm}, clip]{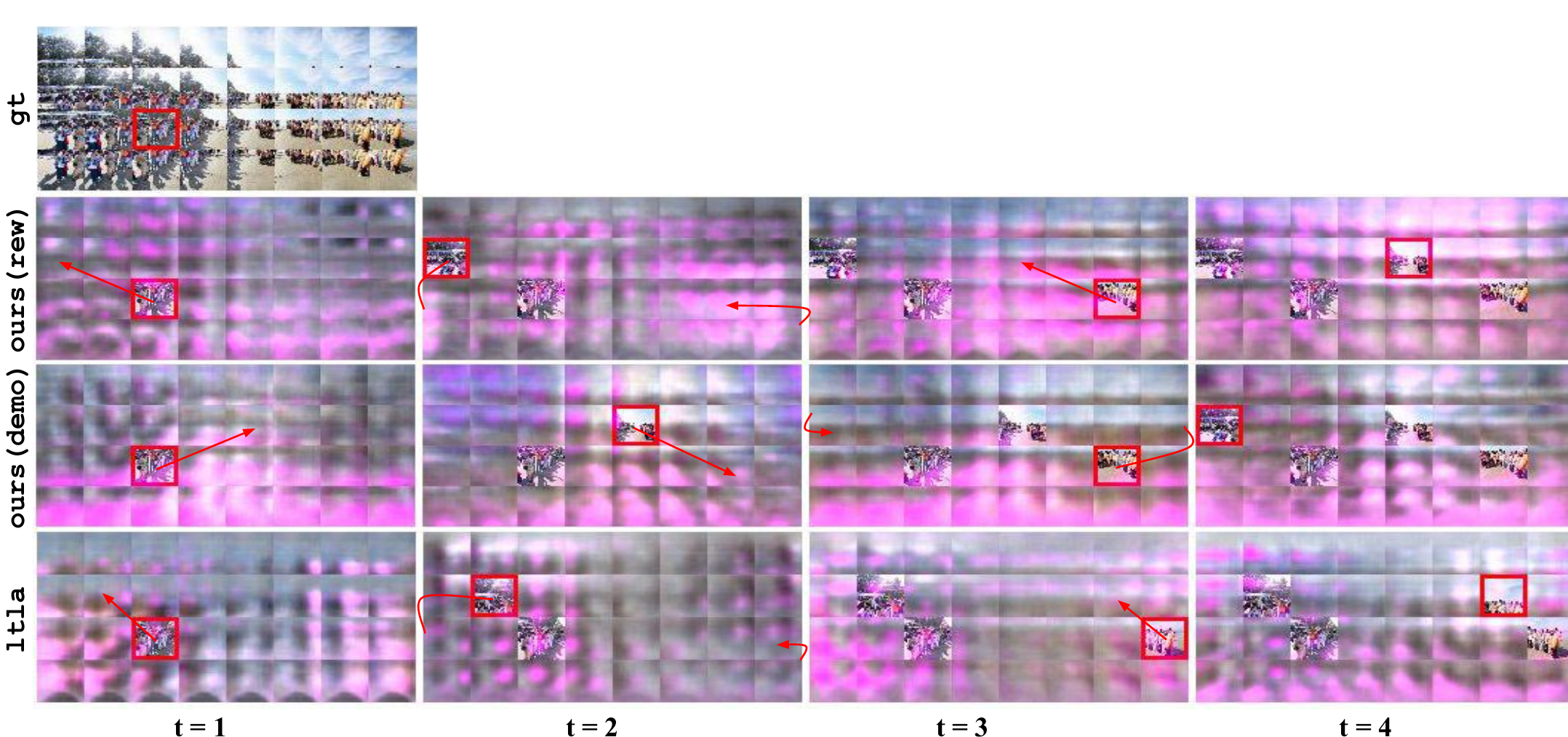}
    \end{subfigure}
    \begin{subfigure}[b]{0.8\textwidth}
        \includegraphics[width=\textwidth, trim={0cm 4cm 3.5cm 0cm}, clip]{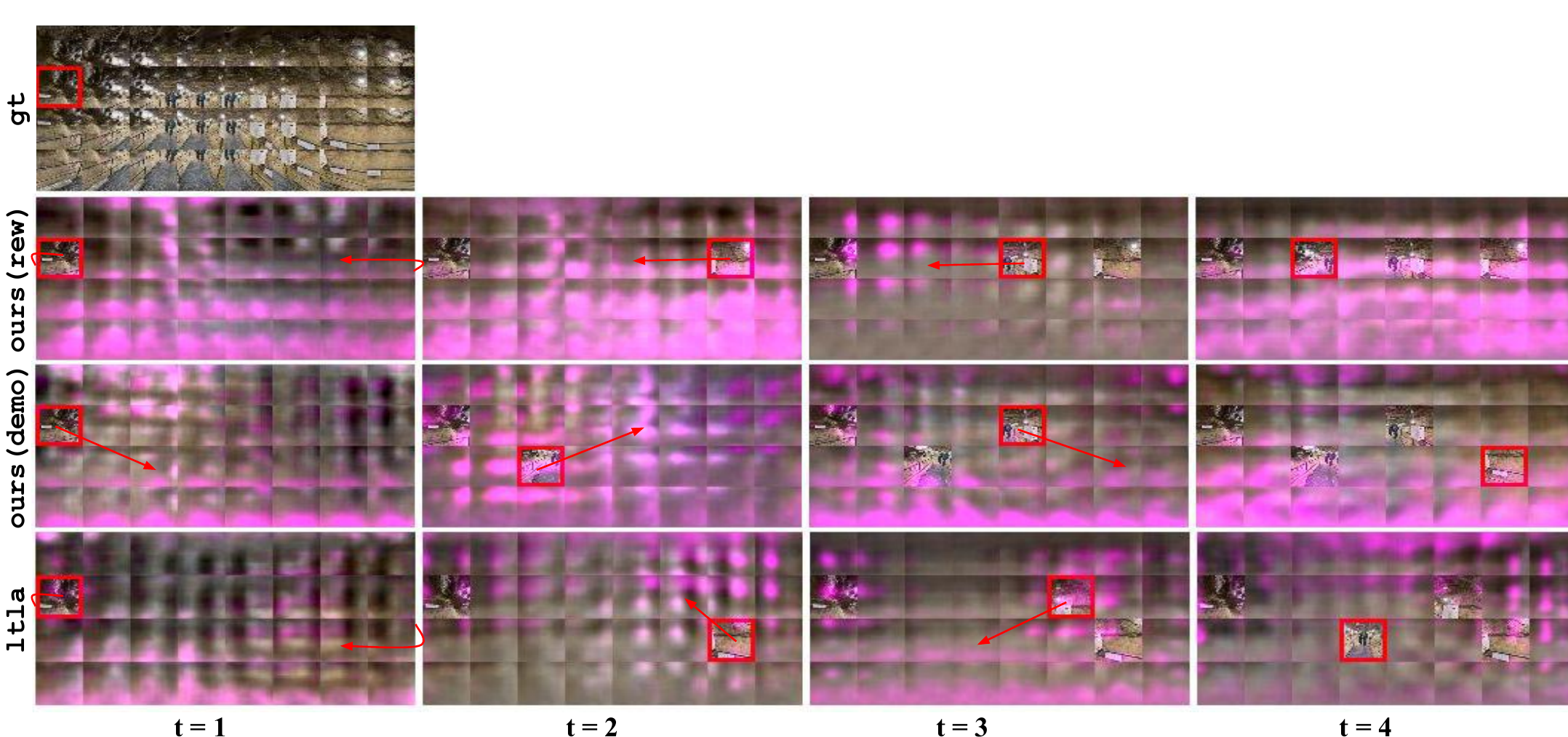}
    \end{subfigure}
    \begin{subfigure}[b]{0.8\textwidth}
        \includegraphics[width=\textwidth, trim={0cm 4cm 3.5cm 0cm}, clip]{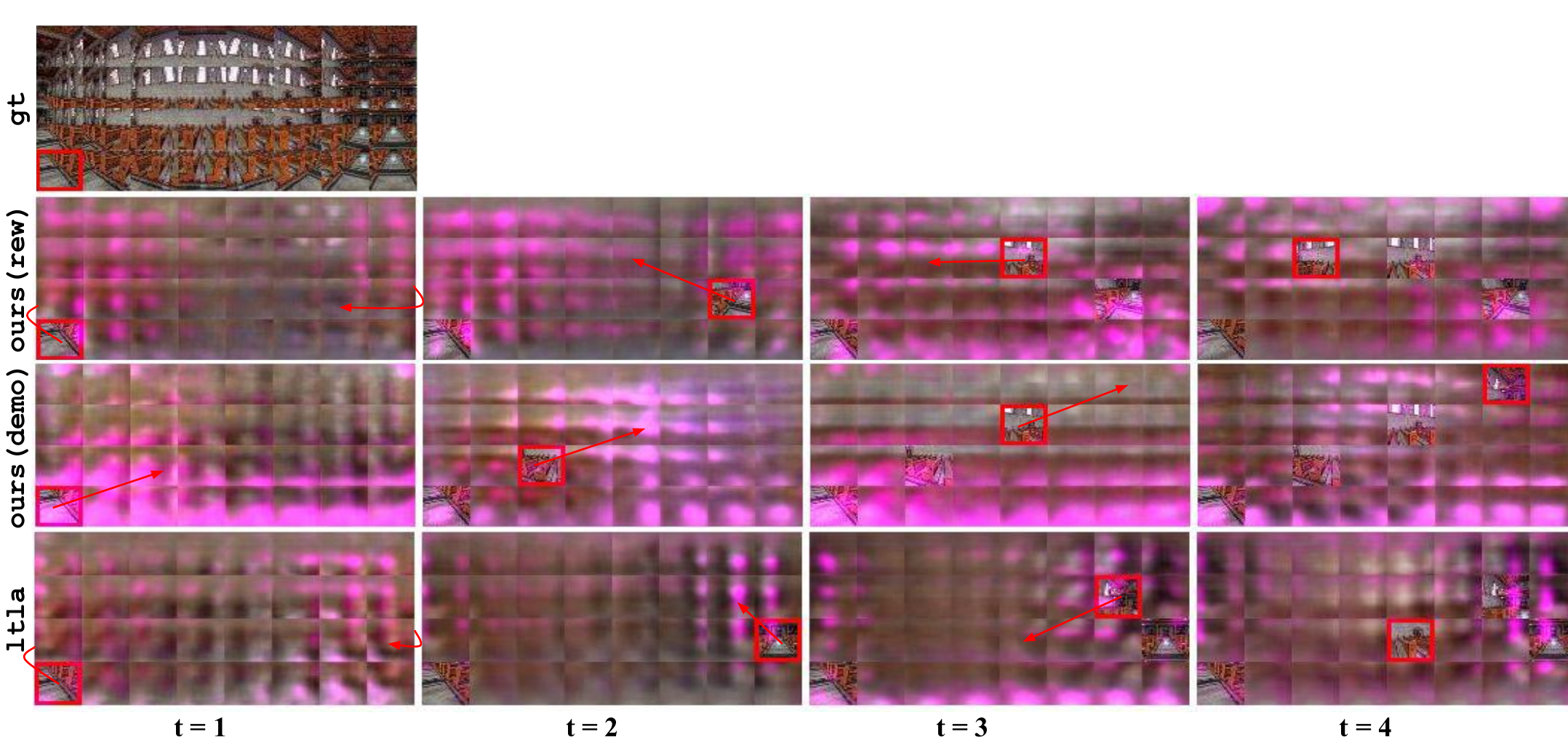}
    \end{subfigure}
    \caption{Policy visualizations of \texttt{ltla}, \texttt{ours(rew)} and \texttt{ours(demo)} on four examples from SUN360. The policies tend to visit regions on the viewgrid with low heatmap densities in order to improve their belief about the environment. Better policies tend to more rapidly improve their beliefs, as witnessed by denser heatmaps. Best viewed on pdf with zoom.}
    \label{fig:sun360_vis}
\end{figure}

Fig.~\ref{fig:sun360_vis} visualizes the policy beliefs for \texttt{ours(demo)}, \texttt{ours(rew)} and \texttt{ltla} on the SUN360 examples from the main paper and an additional example.  Fig.~\ref{fig:modelnet_hard_vis} shows examples for ModelNet Hard.

\begin{figure}
    \centering
    \begin{subfigure}[b]{0.8\textwidth}
        \includegraphics[width=\textwidth, trim={0cm 4cm 3.5cm 0cm}, clip]{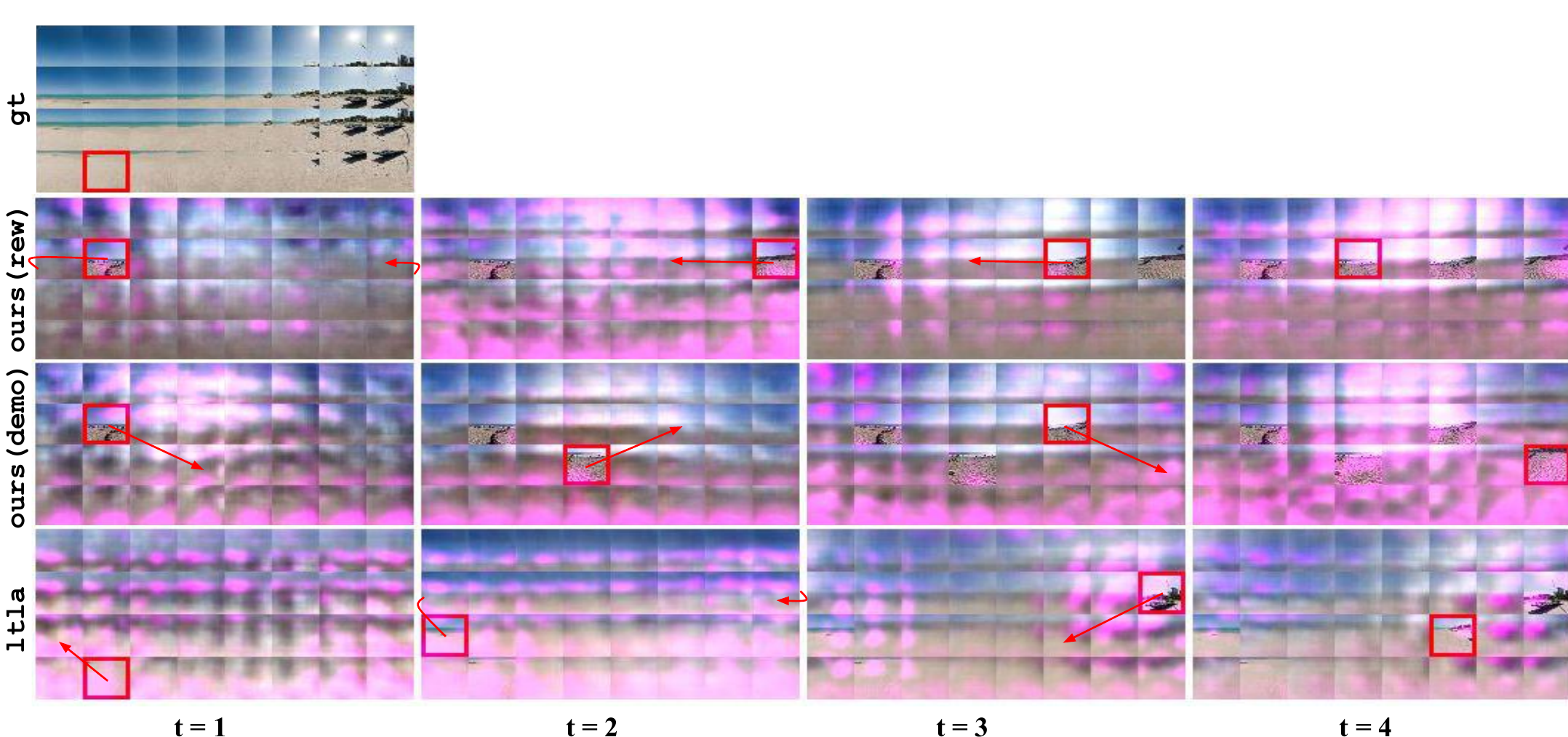}
    \end{subfigure}
    \begin{subfigure}[b]{0.8\textwidth}
        \includegraphics[width=\textwidth, trim={0cm 4cm 3.5cm 0cm}, clip]{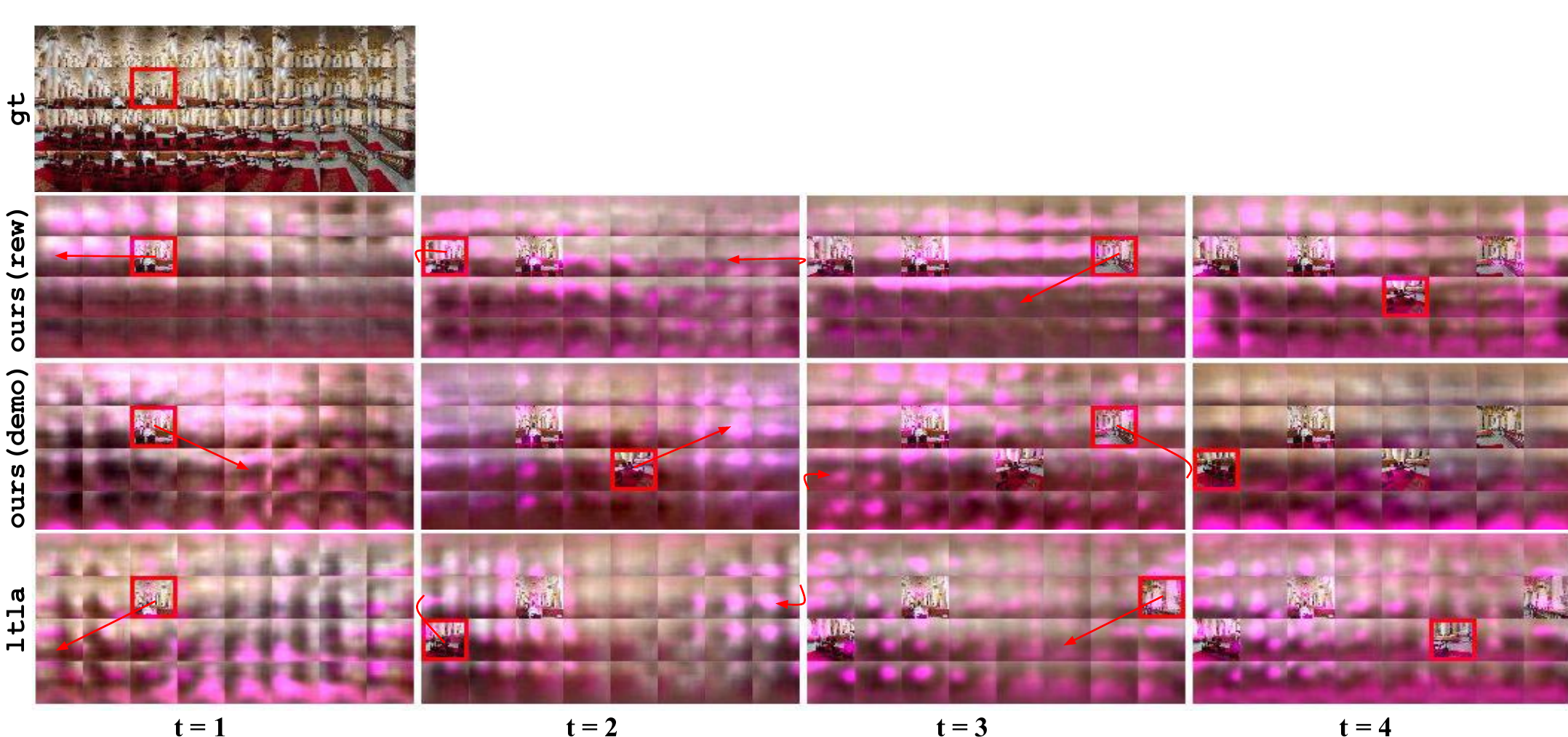}
    \end{subfigure}
    \caption{Examples of less conclusive visualizations on SUN360, where \texttt{ltla}, \texttt{ours(rew)} and \texttt{ours(demo)} have similar heatmap densities.  Best viewed on pdf with zoom.}
    \label{fig:sun360_vis_con}
\end{figure}

\begin{figure}
    \centering
    \begin{subfigure}[b]{\textwidth}
        \includegraphics[width=\textwidth, trim={0cm 4cm 3.5cm 0cm}, clip]{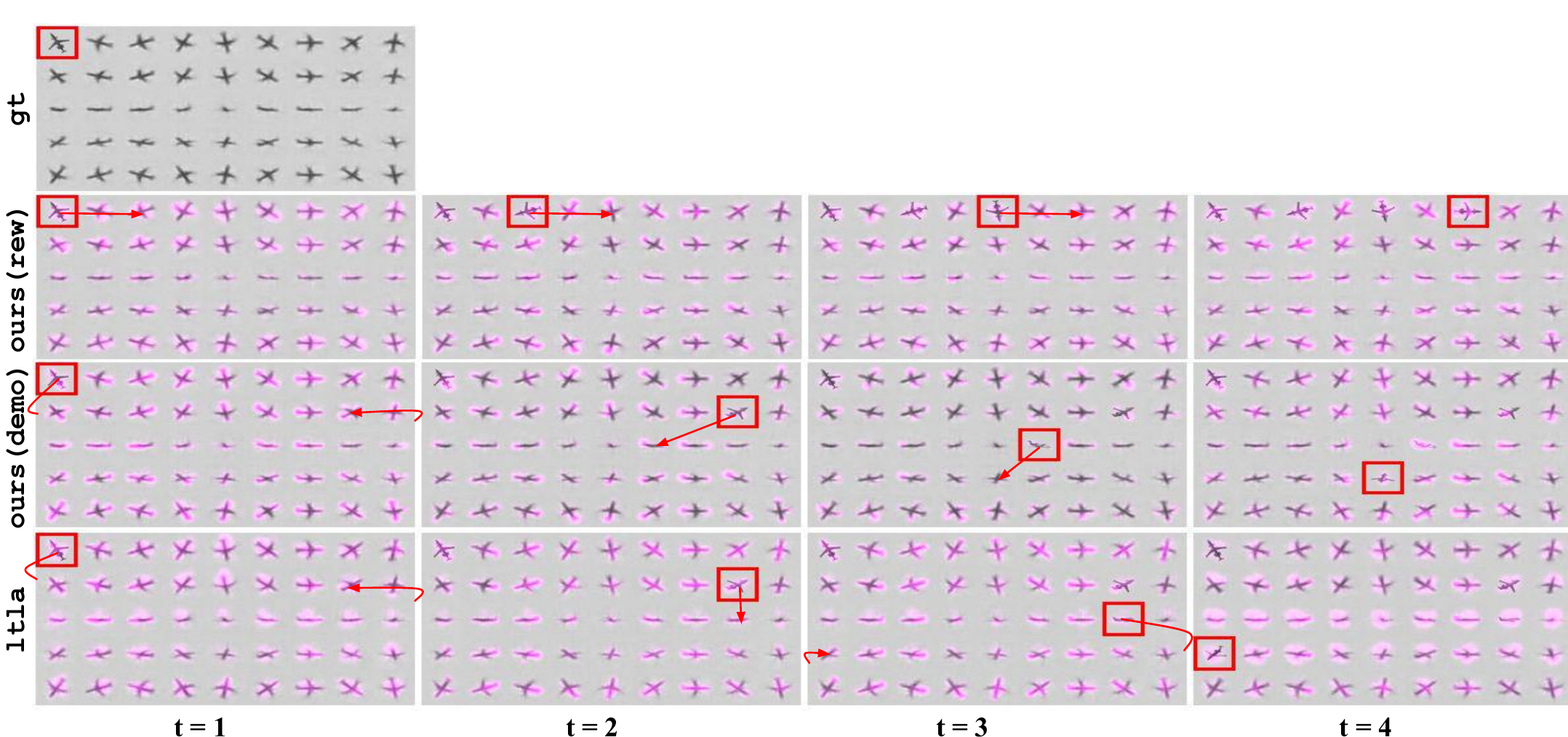}
    \end{subfigure}
    \begin{subfigure}[b]{\textwidth}
        \includegraphics[width=\textwidth, trim={0cm 4cm 3.5cm 0cm}, clip]{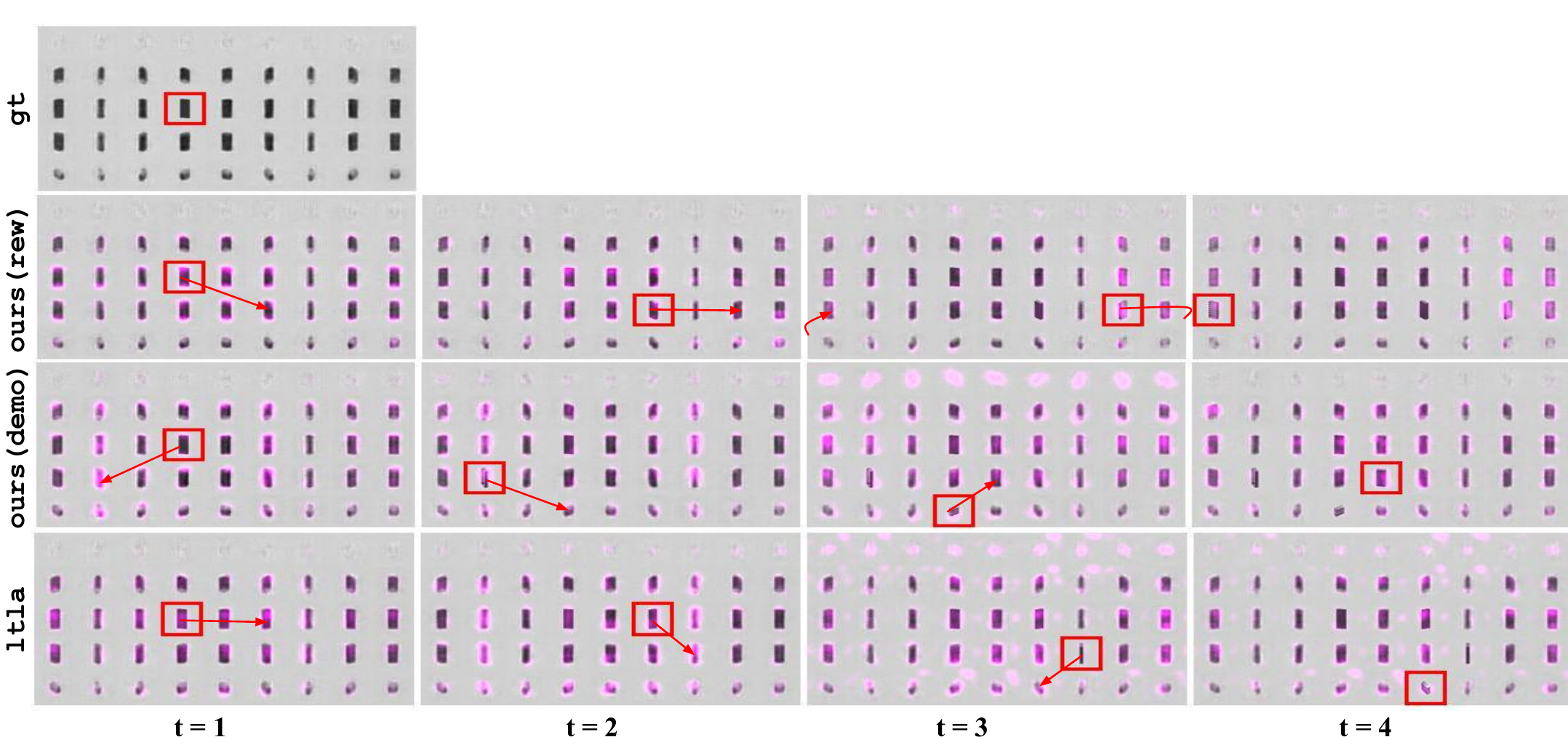}
    \end{subfigure}
    \begin{subfigure}[b]{\textwidth}
        \includegraphics[width=\textwidth, trim={0cm 4cm 3.3cm 0cm}, clip]{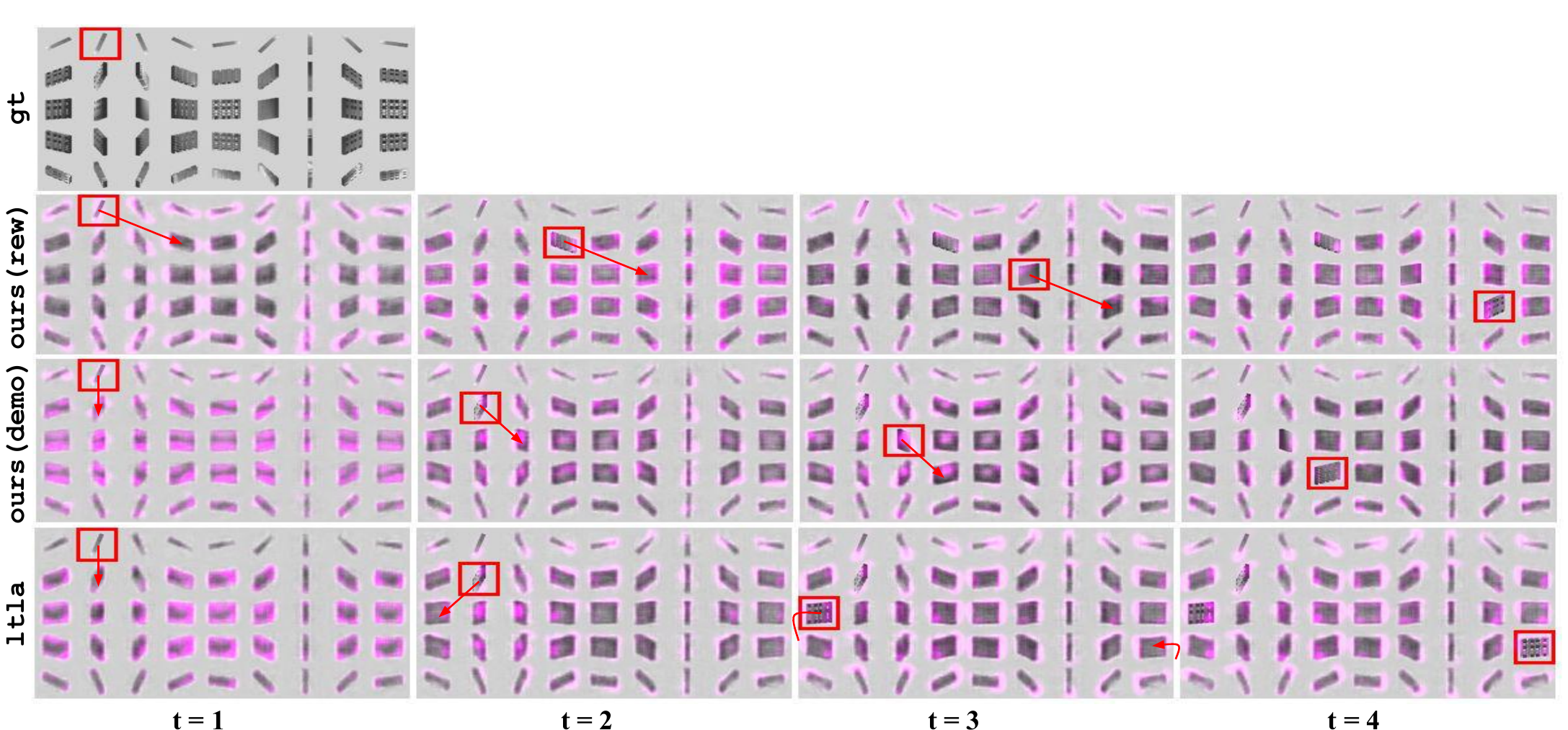}
    \end{subfigure}
    \caption{Policy visualizations of \texttt{ltla} and \texttt{ours(rew)} on three examples from ModelNet Hard.  Best viewed on pdf with zoom.}
    \label{fig:modelnet_hard_vis}
\end{figure}

Fig.~\ref{fig:sun360_vis} shows how all the models follow a
similar behaviour of visiting regions with low heatmap densities, as indicated by the red arrows.
This shows how the agents are often moving towards the views that are not yet contributing effectively to their action selection, to improve their understanding of the scene.
The heatmap ``density'' serves as a high-level visual for the spread of the agent's reasoning about its belief state as it influences its action selection: the greater the spread, the more its belief about the full unobserved environment is directing camera motion selections.
The heatmap density of \texttt{ours(demo)} lies between that of \texttt{ours(rew)} and \texttt{ltla}, which is consistent with the quantitative performance observed (refer to the main paper).  We also note the qualitative difference between the heatmaps of \texttt{ours(demo)} and \texttt{ours(rew)}. While both have dense heatmaps across the entire viewgrid, \texttt{ours(demo)}  appears to rely significantly more on its beliefs about the ground plane of the scene. However, there are cases where the visualizations are not conclusive in differentiating between the policies. As shown in Fig.~\ref{fig:sun360_vis_con}, we can see that visualizations are dense across all models, and therefore, less conclusive.   

In Fig.~\ref{fig:modelnet_hard_vis}, we see our visualization is less effective in differentiating the policies on ModelNet Hard, possibly due to the narrower margins in the reconstruction errors for this dataset.
However, it is interesting to note that the heatmap densities are better concentrated on the object for \texttt{ours(rew)} and \texttt{ours(demo)}, whereas it often unnecessarily leaks to the background pixels for \texttt{ltla}.

\section{Additional training time for sidekicks}
In order to account for additional training time required to train the sidekicks, we analyze the time taken for training various models and sidekicks. Since \emph{all} models are pretrained with $T=1$, including \texttt{ltla} --- the training overhead ($\sim64$ min) is identical for the baseline. Both sidekicks use the $T=1$ model to compute scores (see Sec.~3.4 from main paper), a one-time cost of $\sim 10$ min.  To train for 500 epochs, \texttt{ours (rew)} and \texttt{ours (demo)} require $\sim450$ and $\sim485$ min, resp, while \texttt{ltla} and \texttt{asymm-ac} take $\sim465$ and $\sim529$ min, resp (averaged over 3 runs)~\footnote{Experiments were run on a Intel(R) Xeon(R) CPU @ 1.70GHz system with GeForce GTX 1080 GPU.}. Therefore, the additional training time for sidekicks is nominal in comparison to the overall training process. However, training the expert for \texttt{expert-clone} takes as long as it takes to train a full model ($\sim 1000$ minutes for 1000 epochs), which is $\sim17\times$ the time required to pre-train at $T=1$ and pre-compute the sidekick scores.

\end{document}